\documentclass{article}

\usepackage{graphicx}
\usepackage{amsmath}
\usepackage{amssymb}
\usepackage{booktabs}
\usepackage{color}
\usepackage{natbib} 
\usepackage{multirow}
\usepackage{bm}
\usepackage{amsthm}
\usepackage{url}
\usepackage{subfigure}
\usepackage{algorithm}
\usepackage{algorithmic}
\usepackage{tabulary}

\newtheorem{theorem}{Theorem}[section]
\newtheorem{proposition}[theorem]{Proposition}
\newtheorem{lemma}[theorem]{Lemma}

\theoremstyle{definition}

\newtheorem{assumption}[theorem]{Assumption}
\newtheorem{prob}[theorem]{Problem}
\theoremstyle{remark}

\numberwithin{equation}{section}

\begin{document}
\makeatletter

\begin{center}
\large{\bf Relationship between Batch Size and Number of Steps Needed for Nonconvex Optimization of Stochastic Gradient Descent using Armijo Line Search}\\
\end{center}\vspace{3mm}

\begin{center}
\textsc{Yuki Tsukada}\\ 
Meiji University,
1-1-1 Higashimita, Tama-ku, Kawasaki-shi, Kanagawa 214-8571 Japan.
(yuki.t.1119@iiduka.net)\\
\textsc{Hideaki Iiduka}\\
Department of Computer Science, 
Meiji University,
1-1-1 Higashimita, Tama-ku, Kawasaki-shi, Kanagawa 214-8571 Japan. 
(iiduka@cs.meiji.ac.jp)
\end{center}

\vspace{2mm}

\begin{center}
\begin{minipage}{14cm} 

{\bf Abstract:}
While stochastic gradient descent (SGD) can use various learning rates, such as constant or diminishing rates, the previous numerical results showed that SGD performs better than other deep learning optimizers using when it uses learning rates given by line search methods. 
In this paper, we perform a convergence analysis on SGD with a learning rate given by an Armijo line search for nonconvex optimization
indicating that the upper bound of the expectation of the squared norm of the full gradient becomes small when the number of steps and the batch size are large. 
Next, we show that, for SGD with the Armijo-line-search learning rate, the number of steps needed for nonconvex optimization is a monotone decreasing convex function of the batch size; that is, the number of steps needed for nonconvex optimization decreases as the batch size increases. 
Furthermore, we show that the stochastic first-order oracle (SFO) complexity, which is the stochastic gradient computation cost, is a convex function of the batch size; that is, there exists a critical batch size that minimizes the SFO complexity. 
Finally, we provide numerical results that support our theoretical results. 
The numerical results indicate that the number of steps needed for training deep neural networks decreases as the batch size increases and that there exist the critical batch sizes that can be estimated from the theoretical results.
\end{minipage}
\end{center}

\section{Introduction}
\subsection{Background}
\label{subsec:1.1}
Nonconvex optimization is useful for training deep neural networks, since the loss functions called the expected risk and empirical risk are nonconvex and they need only be minimized in order to find the model parameters. Deep-learning optimizers have been presented for minimizing the loss functions. The simplest one is stochastic gradient descent (SGD) \citep{robb1951,zinkevich2003,nem2009,gha2012,gha2013} and there are numerous theoretical analyses on using SGD for nonconvex optimization \citep{JMLR:v18:16-595,NEURIPS2019_2557911c,feh2020,chen2020,sca2020,loizou2021}. Variants have also been presented, such as momentum methods \citep{polyak1964,nest1983} and adaptive methods including Adaptive Gradient (AdaGrad) \citep{adagrad}, Root Mean Square Propagation (RMSProp) \citep{rmsprop}, Adaptive Moment Estimation (Adam) \citep{adam}, Adaptive Mean Square Gradient (AMSGrad) \citep{reddi2018}, and Adam with decoupled weight decay (AdamW) \citep{loshchilov2018decoupled}. SGD and its variants are useful for training not only deep neural networks but also generative adversarial networks \citep{NIPS2017_8a1d6947,pmlr-v206-naganuma23a,pmlr-v202-sato23b}. 
 
The performance of deep-learning optimizers for nonconvex optimization depends on the batch size. The previous numerical results in \citep{shallue2019} and \citep{zhang2019} have shown that the number of steps $K$ needed to train a deep neural network halves for each doubling of the batch size $b$ and that there is a region of diminishing returns beyond the {\em critical batch size} $b^\star$. This fact can be expressed as follows: there is a positive number $C$ such that $N := K b \approx C$ for $b \leq b^\star$ and $N := K b \geq C$ for $b \geq b^\star$. The deep neural network model uses $b$ gradients of the loss functions per step. Hence, when $K$ is the number of steps required to train a deep neural network, the model has a stochastic gradient computation cost of $Kb$. We will define the {\em stochastic first-order oracle (SFO) complexity} \citep{iiduka2022critical,pmlr-v202-sato23b} of a deep-learning optimizer to be $N := K b$. From the previous numerical results in \citep{shallue2019} and \citep{zhang2019}, the SFO complexity is minimized at a critical batch size $b^\star$ and there are diminishing returns once the batch size exceeds $b^\star$. Therefore, it is desirable to use the critical batch size when minimizing the SFO complexity of the deep-learning optimizer. 

Not only a batch size but also a learning rate affects the performance of deep-learning optimizers for nonconvex optimization. A performance measure of a deep-learning optimizer generating a sequence $(\bm{\theta}_k)_{k\in\mathbb{N}}$ is the expectation of the squared norm of the gradient of a nonconvex loss function $f$, denoted by $\mathbb{E}[\| \nabla f (\bm{\theta}_k)\|^2]$. If this performance measure becomes small when the number of steps $k$ is large, the deep-learning optimizer approximates a local minimizer of $f$. For example, let us consider the problem of minimizing a smooth function $f$ (see Section \ref{subsec:2.1} for the definition of smoothness). Here, SGD using a constant learning rate $\alpha = O(\frac{1}{L})$ satisfies $\min_{k\in [K]} \mathbb{E}\left[\| \nabla f (\bm{\theta}_k)\|^2 \right] = O (\frac{1}{K}+ \frac{\alpha}{b})$, where $L$ is the Lipschitz constant of $\nabla f$, $b$ is the batch size, and $[K] := \{1,2,\ldots,K\}$ (see also Table \ref{table:1}). Moreover, SGD using a learning rate satisfying the Armijo condition was presented in \citep{NEURIPS2019_2557911c}. The {\em Armijo line search} \citep[Chapter 3.1]{noce} is a standard method for finding an appropriate learning rate $\alpha_k$ giving a sufficient decrease in $f$, i.e., $f(\bm{\theta}_{k+1}) < f(\bm{\theta}_k)$ (see Section \ref{subsec:2.3.1} for the definition of the Armijo condition).

\subsection{Motivation}
The numerical results in \citep{NEURIPS2019_2557911c} indicated that the Armijo-line-search learning rate is superior to using a constant learning rate when using SGD to train deep neural networks in the sense of minimizing the training loss functions and improving test accuracy. Motivated by the useful numerical results in \citep{NEURIPS2019_2557911c}, we decided to perform convergence analyses on SGD with the Armijo-line-search learning rate for nonconvex optimization in deep neural networks.

Theorem 3 in \citep{NEURIPS2019_2557911c} is a convergence analysis of SGD with the Armijo-line-search learning rate for nonconvex optimization under a strong growth condition that implies the interpolation property. Here, let $f \colon \mathbb{R}^d \to \mathbb{R}$ be an empirical risk defined by $f(\bm{\theta}) := \frac{1}{n} \sum_{i\in [n]} f_i (\bm{\theta})$, where $n$ is the number of training data and $f_i \colon \mathbb{R}^d \to \mathbb {R}$ is a loss function corresponding to the $i$-th training data $z_i$. We say that $f$ has the interpolation property if $\nabla f(\bm{\theta}) = \bm{0}$ implies $\nabla f_i (\bm{\theta}) = \bm{0}$ $(i\in [n])$. The interpolation property holds for optimization of a linear model with the squared hinge loss for binary classification on linearly separable data \citep[Section 2]{NEURIPS2019_2557911c}. However, the interpolation condition would be unrealistic for deep neural networks, since their loss functions are nonconvex. The motivation behind this work is thus to show that SGD with the Armijo-line-search learning rate can solve nonconvex optimization problems in deep neural networks.

As indicated the second paragraph in Section \ref{subsec:1.1}, the batch size has a significant effect on the performance of SGD. Hence, in accordance with the first motivation stated above, we decided to investigate appropriate batch sizes for SGD with the Armijo-line-search learning rate. In particular, we were interested in verifying whether a critical batch size $b^\star$ minimizing the SFO complexity $N$ exists for training deep neural networks with SGD using the Armijo condition in theory and in practice. This is because the previous studies in \citep{shallue2019,zhang2019,iiduka2022critical,pmlr-v202-sato23b} showed the existence of critical batch sizes for training deep neural networks or generative adversarial networks with optimizers with constant or diminishing learning rates and without Armijo-line-search learning rates.

We are also interested in estimating critical batch sizes before implementing SGD with the Armijo-line-search learning rate. The previous results in \citep{iiduka2022critical,pmlr-v202-sato23b} showed that, for optimizers using constant learning rates, the critical batch sizes determined from numerical results are close to the theoretically estimated sizes. Motivated by the results in \citep{iiduka2022critical,pmlr-v202-sato23b}, we sought to verify whether, for SGD with the Armijo-line-search learning rate, the measured critical batch sizes are close to the batch sizes estimated from theoretical results.

\subsection{Contribution}
\subsubsection{Convergence analysis of SGD with Armijo-line-search learning rates}
\label{subsec:1.3.1}
The first contribution of this paper is to present a convergence analysis of SGD with Armijo-line-search learning rates for general nonconvex optimization (Theorem \ref{thm:1}); in particular, it is shown that SGD with this rate $\alpha_k$ satisfies that, for all $K \geq 1$,
\begin{align}\label{main}
\min_{k\in [0:K-1]} \mathbb{E}\left[ \| \nabla f(\bm{\theta}_k)\|^2 \right]
\leq 
\underbrace{\overbrace{\frac{2(f(\bm{\theta}_{0}) - f_*)}
{\tilde{\alpha} - (L_n \overline{\alpha} -1) \overline{\alpha}}}^{C_1}
\frac{1}{K}}_{B(\bm{\theta}_0, K)} 
+ 
\underbrace{\overbrace{\frac{L_n \overline{\alpha}^2 \sigma^2}{\tilde{\alpha} - (L_n \overline{\alpha} -1) \overline{\alpha}}}^{C_2} \frac{1}{b}}_{V(\sigma^2, b)},
\end{align}
where the parameters are defined in Table \ref{table:1} (see also Theorem \ref{thm:1}). The inequality (\ref{main}) indicates that the upper bound of the performance measure $\min_{k\in [0:K-1]} \mathbb{E}[ \| \nabla f(\bm{\theta}_k)\|^2 ]$ that consists of a bias term $B(\bm{\theta}_0,K)$ and variance term $V(\sigma^2, b)$ becomes small when the number of steps $K$ is large and the batch size $b$ is large. Therefore, it is desirable to set $K$ large and $b$ large so that Algorithm \ref{algo:1} will approximate a local minimizer of $f$.

The essential lemma to proving (\ref{main}) is the guarantee of the existence of a lower bound on the learning rates satisfying the Armijo condition (Lemma \ref{lem:1}). Although, in general, learning rates satisfying the Armijo condition do not have any lower bound (Lemma \ref{lem:1}(i)), the corresponding learning rates computed by a backtracking line search (Algorithm \ref{algo:2}) have a lower bound (Lemma \ref{lem:1}(ii)). In addition, the descent lemma (i.e., $f(\bm{y}) \leq f(\bm{x}) + \langle \nabla f(\bm{x}), \bm{y}-\bm{x} \rangle + \frac{L_n}{2} \|\bm{y}-\bm{x}\|^2$ $(\bm{x},\bm{y} \in \mathbb{R}^d)$) holds from the smoothness condition on $f$. Thus, we can prove (\ref{main}) by using the existence of a lower bound on the learning rates satisfying the Armijo condition and the descent lemma (see Appendix \ref{a2} for details of the proof of Theorem \ref{thm:1}). 

\begin{table*}[ht]
\caption{Relationship between batch size $b$ and number of steps $K$ to achieve an $\epsilon$--approximation defined by $\min_{k\in [0:K-1]} \mathbb{E}[\| \nabla f (\bm{\theta}_k)\|^2 ] \leq \frac{C_1}{K} + \frac{C_2}{b} = \epsilon^2$ for SGD with a constant learning rate $\alpha \in (0,\frac{2}{L_n})$ and for SGD with the Armijo-line-search learning rate $\alpha_k \in [\underline{\alpha}, 
\overline{\alpha}]$ ($[0:K-1] := \{0,1,\ldots,K-1\}$, $f := \frac{1}{n} \sum_{i\in[n]} f_i$ is bounded below by $f_*$, $L_i$ is the Lipschitz constant of $\nabla f_i$, $L_n := \frac{1}{n} \sum_{i\in [n]} L_i$,
$\tilde{\alpha}:=\frac{2 \delta (1-c)}{L_n}$,
$\delta \in (\frac{1}{4},1)$, $c \in (0,1-\frac{1}{4\delta})$,
and   
$\sigma^2$ is the upper bound of the variance of the stochastic gradient)}
\label{table:1}
\centering
\begin{tabular}{llllll}
\toprule
Learning Rate 
& \multicolumn{2}{l}{Upper Bound $\frac{C_1}{K} + \frac{C_2}{b}$} 
& Steps $K$ 
& SFO $N$ 
& Critical Batch $b^\star$ \\
\midrule
\multirow{2}{*}{Constant $\displaystyle{\alpha \in \left(0,\frac{2}{L_n} \right)}$} 
& $C_1$ & $\displaystyle{\frac{2(f(\bm{\theta}_0) - f_*)}{(2-L_n \alpha)\alpha}}$ & \multirow{2}{*}{$\displaystyle{K = \frac{C_1 b}{\epsilon^2 b - C_2}}$} & \multirow{2}{*}{$\displaystyle{N = \frac{C_1 b^2}{\epsilon^2 b - C_2}}$} & \multirow{2}{*}{$\displaystyle{b^\star = \frac{2 C_2}{\epsilon^2}}$} \\
& $C_2$ & $\displaystyle{\frac{L_n \alpha \sigma^2}{2-L_n \alpha}}$ & & & \\
\midrule
\multirow{2}{*}{Armijo $\displaystyle{\overline{\alpha} \in \left(\frac{1}{L_n}, \frac{2}{L_n}\right)}$} 
& $C_1$ & $\displaystyle{\frac{2(f(\bm{\theta}_{0}) - f_*)}
{\tilde{\alpha}- (L_n \overline{\alpha} -1) \overline{\alpha}}}$ & \multirow{2}{*}{$\displaystyle{K = \frac{C_1 b}{\epsilon^2 b - C_2}}$} & \multirow{2}{*}{$\displaystyle{N = \frac{C_1 b^2}{\epsilon^2 b - C_2}}$} & \multirow{2}{*}{$\displaystyle{b^\star = \frac{2 C_2}{\epsilon^2}}$} \\
& $C_2$ & $\displaystyle{\frac{L_n \overline{\alpha}^2 \sigma^2}
{\tilde{\alpha} - (L_n \overline{\alpha} -1) \overline{\alpha}}}$ & & & \\
\bottomrule 
\end{tabular}
\end{table*}


\subsubsection{Steps needed for $\epsilon$--approximation of SGD with Armijo line-search-learning rates}
\label{subsec:1.3.2}
The previous results in \citep{shallue2019,zhang2019,iiduka2022critical,pmlr-v202-sato23b} indicated that, for optimizers, the number of steps $K$ needed to train a deep neural network or generative adversarial networks decreases as the batch size increases. The second contribution of this paper is to show that, for SGD with the Armijo-line-search learning rate, the number of steps $K$ needed for nonconvex optimization decreases as the batch size increases. Let us consider the case in which the right-hand side of (\ref{main}) is equal to $\epsilon^2$, where $\epsilon > 0$ is the precision. Then, $K$ is a rational function defined for a batch size $b$ by 
\begin{align}\label{f_K}
K = K(b) = \frac{C_1 b}{\epsilon^2 b - C_2}, 
\end{align}
where $C_1$ and $C_2$ are the positive constants defined in (\ref{main}) (see also Table \ref{table:1}). We can easily show that $K$ defined above is a monotone decreasing and convex function with respect to $b$ (Theorem \ref{thm:2}). Accordingly, the number of steps needed for nonconvex optimization decreases as the batch size increases.

\subsubsection{Critical batch size minimizing SFO complexity of SGD with Armijo-line-search learning rates}
\label{subsec:1.3.3}
Using $K$ defined by (\ref{f_K}) above, we can further define the SFO complexity $N$ of SGD with Armijo-line-search learning rates (see also Table \ref{table:1}):
\begin{align}\label{f_N}
N = Kb = K(b)b = \frac{C_1 b^2}{\epsilon^2 b - C_2}.
\end{align}
We can easily show that $N$ is convex with respect to $b$ and that a global minimizer 
\begin{align}\label{CBS}
b^\star = \frac{2 C_2}{\epsilon^2} = 
\frac{2 L_n \overline{\alpha}^2 \sigma^2}
{\{ \tilde{\alpha}- (L_n \overline{\alpha} -1) \overline{\alpha}\} \epsilon^2}
\end{align}
exists for it (Theorem \ref{thm:3}). Accordingly, there is a critical batch size $b^\star$ at which $N$ is minimized.

Here, we compare the number of steps $K_{\mathrm{C}}$ and the SFO complexity $N_{\mathrm{C}}$ for SGD using a constant learning rate $\alpha$ with $K_{\mathrm{A}}$ and $N_{\mathrm{A}}$ for SGD using the Armijo-line-search learning rate $\alpha_k$ $(\in [\underline{\alpha}, \overline{\alpha}])$. Let $C_{1,\mathrm{C}}$ (resp. $C_{2,\mathrm{C}}$) be $C_1$ (resp. $C_{2}$) in Table \ref{table:1} for SGD using a constant learning rate and let $C_{1,\mathrm{A}}$ (resp. $C_{2,\mathrm{A}}$) be $C_1$ (resp. $C_2$) in Table \ref{table:1} for SGD using the Armijo-line-search learning rate. We have that
\begin{align}\label{comparison_C}
\begin{split}
&C_{1,\mathrm{A}} < C_{1,\mathrm{C}}
\text{ iff } 
(2-L_n \alpha)\alpha < \tilde{\alpha}- (L_n \overline{\alpha} - 1)\overline{\alpha}, \\
&C_{2,\mathrm{A}} < C_{2,\mathrm{C}}
\text{ iff } 
\frac{\overline{\alpha}^2 \sigma_{\mathrm{A}}^2}{\tilde{\alpha}- (L_n \overline{\alpha} -1) \overline{\alpha}} 
< \frac{\alpha \sigma_{\mathrm{C}}^2}{2 - L_n \alpha},
\end{split}
\end{align} 
where $\sigma_{\mathrm{C}}^2$ (resp. $\sigma_{\mathrm{A}}^2$) denotes the upper bound of the variance of the stochastic gradient for SGD using a constant learning rate $\alpha$ (resp. the Armijo-line-search learning rate). If (\ref{comparison_C}) holds, then SGD using the Armijo-line-search learning rate converges faster than SGD using a constant learning rate in the sense that 
$
\frac{C_{1,\mathrm{A}} b}{\epsilon^2 b - C_{2,\mathrm{A}}} = K_{\mathrm{A}} < K_{\mathrm{C}} = \frac{C_{1,\mathrm{C}} b}{\epsilon^2 b - C_{2,\mathrm{C}}} 
\text{ and } 
\frac{C_{1,\mathrm{A}} b^2}{\epsilon^2 b - C_{2,\mathrm{A}}} = N_{\mathrm{A}} < N_{\mathrm{C}}
= \frac{C_{1,\mathrm{C}} b^2}{\epsilon^2 b - C_{2,\mathrm{C}}}.
$
It would be difficult to check exactly that (\ref{comparison_C}) holds before implementing SGD, since (\ref{comparison_C}) involves unknown parameters, such as $L_n = \frac{1}{n} \sum_{i\in [n]} L_i$, $\sigma_{\mathrm{C}}^2$, and $\sigma_{\mathrm{A}}^2$. However, it can be expected that (\ref{comparison_C}) holds, since it is known empirically \citep[Figure 5]{NEURIPS2019_2557911c} that the relationship between the Armijo-line-search learning rate $\alpha_k$ and a constant learning rate $\alpha$ is 
$\alpha < \alpha_k < \overline{\alpha}$. 

\subsubsection{Numerical results supporting our theoretical results}
The numerical results in \citep{NEURIPS2019_2557911c} showed that SGD with the Armijo-line-search learning rate performs better than other optimizers in training deep neural networks. Hence, we sought to verify whether the numerical results match our theoretical results (Sections \ref{subsec:1.3.1}, \ref{subsec:1.3.2}, and \ref{subsec:1.3.3}). We trained residual networks (ResNets) on the CIFAR-10 and CIFAR-100 datasets and a two-hidden-layer multi-layer perceptron (MLP) on the MNIST dataset. We numerically found that increasing the batch size $b$ decreases the number of steps $K$ needed to achieve high training accuracies and that there are critical batch sizes minimizing the SFO complexities. We also estimated batch sizes using (\ref{CBS}) for the critical batch size $b^\star$ and compared them with ones determined from the numerical results. We found that the estimated batch sizes are close to the ones determined from the numerical results. To verify whether SGD using the Armijo-line-search learning rate performs better than SGD using a constant learning rate (see the discussion in condition (\ref{comparison_C})), we numerically compared SGD using the Armijo-line-search learning rate with not only SGD using a constant learning rate but also variants of SGD, such as the momentum method, Adam, AdamW, and RMSProp. We found that SGD using the Armijo-line-search learning rate and the critical batch size performs better than other optimizers in the sense of minimizing the number of steps and the SFO complexities needed to achieve high training accuracies (Section \ref{sec:4}).

\section{Mathematical Preliminaries}
\subsection{Definitions}
\label{subsec:2.1}
Let $\mathbb{N}$ be the set of nonnegative integers, $[n] := \{1,2,\ldots,n\}$ for $n \geq 1$, and $[0:n] := \{0,1,\ldots,n\}$ for $n \geq 0$. Let $\mathbb{R}^d$ be a $d$--dimensional Euclidean space with inner product $\langle \cdot, \cdot \rangle$ inducing the norm $\|\cdot\|$. 

Let $f \colon \mathbb{R}^d \to \mathbb{R}$ be continuously differentiable. We denote the gradient of $f$ by $\nabla f \colon \mathbb{R}^d \to \mathbb{R}^d$. Let $L>0$. $f \colon \mathbb{R}^d \to \mathbb{R}$ is said to be $L$--smooth if $\nabla f \colon \mathbb{R}^d \to \mathbb{R}^d$ is $L$--Lipschitz continuous, i.e., for all $\bm{x},\bm{y} \in \mathbb{R}^d$, $\|\nabla f(\bm{x}) - \nabla f(\bm{y}) \| \leq L \|\bm{x}-\bm{y}\|$. When $f \colon \mathbb{R}^d \to \mathbb{R}$ is $L$--smooth, the following inequality, called the descent lemma \citep[Lemma 5.7]{doi:10.1137/1.9781611974997}, holds: for all $\bm{x},\bm{y} \in \mathbb{R}^d$, $f(\bm{y}) \leq f(\bm{x}) + \langle \nabla f(\bm{x}), \bm{y}-\bm{x} \rangle + \frac{L}{2} \|\bm{y}-\bm{x}\|^2$. Let $f_* \in \mathbb{R}$. $f \colon \mathbb{R}^d \to \mathbb{R}$ is said to be bounded below by $f_*$ if, for all $\bm{x} \in \mathbb{R}^d$, $f(\bm{x}) \geq f_*$.

\subsection{Assumptions and problem}
Given a parameter $\bm{\theta} \in \mathbb{R}^d$ and a data point $z$ in a data domain $Z$, a machine learning model provides a prediction whose quality can be measured by a differentiable nonconvex loss function $f(\bm{\theta};z)$. We aim to minimize the empirical average loss defined for all $\bm{\theta} \in \mathbb{R}^d$ by 
$
f(\bm{\theta}) = \frac{1}{n} \sum_{i\in [n]} f(\bm{\theta};z_i)
= \frac{1}{n} \sum_{i\in [n]} f_i(\bm{\theta}),
$
where $S = (z_1, z_2, \ldots, z_n)$ denotes the training set and $f_i (\cdot) := f(\cdot;z_i)$ denotes the loss function corresponding to the $i$-th training data $z_i$. 

This paper considers the following smooth nonconvex optimization problem.

\begin{prob}\label{prob:1}
Suppose that $f_i \colon \mathbb{R}^d \to \mathbb{R}$ $(i\in [n])$ is $L_i$--smooth and bounded below by $f_{i,*}$. Then, 
\begin{align*}
\text{minimize } f(\bm{\theta}) := \frac{1}{n} \sum_{i\in [n]} f_i (\bm{\theta})
\text{ subject to } \bm{\theta} \in \mathbb{R}^d.
\end{align*}
\end{prob}

We assume that a stochastic first-order oracle (SFO) exists such that, for a given $\bm{\theta} \in \mathbb{R}^d$, it returns a stochastic gradient $\mathsf{G}_{\xi}(\bm{\theta})$ of the function $f$, where a random variable $\xi$ is supported on a finite/an infinite set $\Xi$ 
independently of $\bm{\theta}$. We make the following standard assumptions.

\begin{assumption}
\label{assum:1}
\text{}
\begin{enumerate} 
\item[(A1)] Let $(\bm{\theta}_k)_{k\in \mathbb{N}} \subset \mathbb{R}^d$ be the sequence generated by SGD. For each iteration $k$, 
\begin{align}\label{gradient}
\mathbb{E}_{\xi_k}\left[ \mathsf{G}_{\xi_k}(\bm{\theta}_k) \right] = \nabla f(\bm{\theta}_k),
\end{align}
where $\xi_0, \xi_1, \ldots$ are independent samples, the random variable $\xi_k$ is independent of $(\bm{\theta}_l)_{l=0}^k$, and 
$\mathbb{E}_{\xi_k}[\cdot]$ stands for the expectation with respect to $\xi_k$.
There exists a nonnegative constant $\sigma^2$ such that 
\begin{align}\label{sigma}
\mathbb{E}_{\xi_k} \left[ \|\mathsf{G}_{\xi_k}(\bm{\theta}_k) - 
\nabla f(\bm{\theta}_k) \|^2 \right] \leq \sigma^2.
\end{align}
\item[(A2)] For each iteration $k$, SGD samples a batch $B_{k}$ of size $b$ independently of $k$ and estimates the full gradient $\nabla f$ as 
\begin{align*}
\nabla f_{B_k} (\bm{\theta}_k)
:= \frac{1}{b} \sum_{i\in [b]} \mathsf{G}_{\xi_{k,i}}(\bm{\theta}_k)
= \frac{1}{b} \sum_{i\in [b]} \nabla f_{\xi_{k,i}}(\bm{\theta}_k),
\end{align*}
where $\xi_{k,i}$ is a random variable generated by the $i$-th sampling in the $k$-th iteration. 
\end{enumerate}
\end{assumption}
From the independence of $\xi_0, \xi_1, \ldots$, we can define the total expectation $\mathbb{E}$ by 
$\mathbb{E} = \mathbb{E}_{\xi_0} \mathbb{E}_{\xi_1} \cdots \mathbb{E}_{\xi_k}$ 
.

\subsection{Stochastic gradient descent using Armijo line search}
\subsubsection{Armijo condition}\label{subsec:2.3.1}
Suppose that $f \colon \mathbb{R}^d \to \mathbb{R}$ is continuously differentiable. We would like to find a stationary point $\bm{\theta}^\star \in \mathbb{R}^d$ such that $\nabla f (\bm{\theta}^\star) = \bm{0}$ by using an iterative method defined by 
\begin{align}\label{iterative}
\bm{\theta}_{k+1} := \bm{\theta}_{k} + \alpha_k \bm{d}_k,
\end{align}
where $\alpha_k > 0$ is the step size (called a learning rate in the machine learning field) and $\bm{d}_k \in \mathbb{R}^d$ is the search direction. Various methods can be used depending on the search direction $\bm{d}_k$. For example, the method (\ref{iterative}) with $\bm{d}_k := - \nabla f(\bm{\theta}_k)$ is gradient descent, while the method (\ref{iterative}) with $\bm{d}_k := - \nabla f(\bm{\theta}_k) + \beta_{k-1} \bm{d}_{k-1}$, where $\beta_k \geq 0$, is the conjugate gradient method. If we define $\bm{d}_k$ (e.g., $\bm{d}_k := - \nabla f(\bm{\theta}_k)$), it is desirable to set $\alpha_k^\star$ satisfying 
\begin{align}\label{optimal_ss}
f(\bm{\theta}_{k} + \alpha_k^\star \bm{d}_k) 
= \min_{\alpha > 0} f(\bm{\theta}_{k} + \alpha \bm{d}_k).
\end{align}
The step size $\alpha_k^\star$ defined by (\ref{optimal_ss}) can be easily computed when $f$ is quadratic and convex. However, for a general nonconvex function $f$, it is difficult to compute the step size $\alpha_k^\star$ in (\ref{optimal_ss}) exactly. Here, we can use the {\em Armijo condition} for finding an appropriate step size $\alpha_k$: Let $c \in (0,1)$. We would like to find $\alpha_k > 0$ such that 
\begin{align}\label{armijo_0}
f(\bm{\theta}_k + \alpha_k \bm{d}_k)
\leq 
f(\bm{\theta}_k) + c \alpha_k \langle \nabla f(\bm{\theta}_k), \bm{d}_k \rangle.
\end{align}
When $\bm{d}_k$ satisfies the descent property defined by $\langle \nabla f(\bm{\theta}_k), \bm{d}_k \rangle < 0$ (e.g., gradient descent using $\bm{d}_k := - \nabla f(\bm{\theta}_k)$ has the property such that $\langle \nabla f(\bm{\theta}_k), \bm{d}_k \rangle = - \|\nabla f(\bm{\theta}_k)\|^2 < 0$), the Armijo condition ensures that $f(\bm{\theta}_{k+1}) = f(\bm{\theta}_k + \alpha_k \bm{d}_k) < f(\bm{\theta}_k)$. Accordingly, $\alpha_k$ satisfying the Armijo condition (\ref{armijo_0}) is appropriate in the sense of minimizing $f$.

The existence of step sizes satisfying the Armijo condition (\ref{armijo_0}) is guaranteed. 

\begin{proposition}{\em \citep[Lemma 3.1]{noce}}\label{existence_armijo}
Let $f \colon \mathbb{R}^d \to \mathbb{R}$ be continuously differentiable. Let $\bm{\theta}_k \in \mathbb{R}^d$ and let $\bm{d}_k$ $(\neq \bm{0})$ have the descent property defined by $\langle \nabla f(\bm{\theta}_k), \bm{d}_k \rangle < 0$. Let $c \in (0,1)$. Then, there exists $\gamma_k > 0$ such that, for all $\alpha_k \in (0,\gamma_k]$, the Armijo condition (\ref{armijo_0}) holds.
\end{proposition}

\subsubsection{Stochastic gradient descent under Armijo condition} 
The objective of this paper is to solve Problem \ref{prob:1} using mini-batch SGD under Assumption \ref{assum:1} defined by 
\begin{align*}
\bm{\theta}_{k+1}
= \bm{\theta}_k + \alpha_k \bm{d}_k
= \bm{\theta}_k - \alpha_k \nabla f_{B_k}(\bm{\theta}_k)
= \bm{\theta}_k - \frac{\alpha_k}{b} \sum_{i \in [b]} \mathsf{G}_{\xi_{k,i}} (\bm{\theta}_k), 
\end{align*}

where $b > 0$ is the batch size and $\alpha_k > 0$ is the learning rate. For each iteration $k$, we can use $\bm{\theta}_k$, $f_{B_k}$, and $\nabla f_{B_k}$. Hence, the Armijo condition \citep[(1)]{NEURIPS2019_2557911c} at the $k$-th iteration for SGD can be obtained by replacing $f$ in (\ref{armijo_0}) with $f_{B_k}$ and using $\bm{d}_k = - \nabla f_{B_k}(\bm{\theta}_k)$:
\begin{align}\label{armijo}
f_{B_k}(\bm{\theta}_k - \alpha_k \nabla f_{B_k}(\bm{\theta}_k))
\leq 
f_{B_k}(\bm{\theta}_k) - c \alpha_k \|\nabla f_{B_k}(\bm{\theta}_k)\|^2.
\end{align}
The Armijo condition (\ref{armijo}) ensures that $f_{B_k}(\bm{\theta}_{k+1}) = f_{B_k}(\bm{\theta}_k - \alpha_k \nabla f_{B_k}(\bm{\theta}_k)) < f_{B_k}(\bm{\theta}_k)$; i.e., the Armijo condition (\ref{armijo}) is appropriate in the sense of minimizing the estimated objective function $f_{B_k}$ from the full objective function $f$. In fact, the numerical results in \citep[Section 7]{NEURIPS2019_2557911c} indicate that SGD using the Armijo condition (\ref{armijo}) is superior to using other deep-learning optimizers to train deep neural networks.

Algorithm \ref{algo:1} is the SGD algorithm using the Armijo condition (\ref{armijo}).

\begin{algorithm}
\caption{Stochastic gradient descent using Armijo line search}
\label{algo:1}
\begin{algorithmic}
\REQUIRE
$c \in (0,1)$ (hyperparameter), $b > 0$ (batch size), $\bm{\theta}_0 \in \mathbb{R}^d$ (initial point), $K \geq 1$ (steps)
\ENSURE 
$\bm{\theta}_K \in \mathbb{R}^d$
\STATE{$k \gets 0$}
\FOR{$k=0,1,\ldots,K-1$}
\STATE{Compute $\alpha_k > 0$ satisfying
$f_{B_k}(\bm{\theta}_k - \alpha_k \nabla f_{B_k}(\bm{\theta}_k))
\leq 
f_{B_k}(\bm{\theta}_k) - c \alpha_k \|\nabla f_{B_k}(\bm{\theta}_k)\|^2$ $\triangleleft$ Algorithm \ref{algo:2}}
\STATE{Compute $\bm{\theta}_{k+1} = \bm{\theta}_k - \alpha_k \nabla f_{B_k}(\bm{\theta}_k)$}
\ENDFOR
\end{algorithmic}
\end{algorithm}

The search direction of Algorithm \ref{algo:1} is $\bm{d}_k = - \nabla f_{B_k}(\bm{\theta}_k)$ $(\neq \bm{0})$ which has the descent property defined by $\langle \nabla f_{B_k}(\bm{\theta}_k), \bm{d}_k \rangle = - \|\nabla f_{B_k}(\bm{\theta}_k)\|^2 < 0$. Hence, from Proposition \ref{existence_armijo}, there exists a learning rate $\alpha_k \in (0,\gamma_k]$ satisfying the Armijo condition (\ref{armijo}). Moreover, the proposition guarantees that the learning rate can be chosen to be sufficiently small, e.g., $\liminf_{k \to + \infty} \alpha_k = 0$.

The convergence analyses of Algorithm \ref{algo:1} use a lower bound of $\alpha_k \in (0,\gamma_k]$ satisfying the Armijo condition (\ref{armijo}). To guarantee the existence of such a lower bound, we use the backtracking method (\citep[Algorithm 3.1]{noce} and \citep[Algorithm 2]{NEURIPS2019_2557911c}) described in Algorithm \ref{algo:2}.

\begin{algorithm}
\caption{Backtracking Armijo-line-search method \citep[Algorithm 3.1]{noce}}
\label{algo:2}
\begin{algorithmic}
\REQUIRE
$c, \delta, \frac{1}{\gamma} \in (0,1)$ (hyperparameters), $\alpha = \gamma^{\frac{b}{n}} \alpha_{k-1}$ (initialization), 
$\bm{\theta}_k \in \mathbb{R}^d$, $f_{B_k} \colon \mathbb{R}^d \to \mathbb{R}$
\ENSURE
$\alpha_k$ satisfying $f_{B_k}(\bm{\theta}_k - \alpha_k \nabla f_{B_k}(\bm{\theta}_k))
\leq 
f_{B_k}(\bm{\theta}_k) - c \alpha_k \|\nabla f_{B_k}(\bm{\theta}_k)\|^2$
\REPEAT
\STATE{$\alpha \gets \delta \alpha$}
\UNTIL{$f_{B_k}(\bm{\theta}_k - \alpha \nabla f_{B_k}(\bm{\theta}_k))
\leq 
f_{B_k}(\bm{\theta}_k) - c \alpha \|\nabla f_{B_k}(\bm{\theta}_k)\|^2$}
\end{algorithmic}
\end{algorithm}

The following lemma guarantees the existence of a lower bound on the learning rates computed by Algorithm \ref{algo:2}. The proof is given in Appendix \ref{a1}.

\begin{lemma}\label{lem:1}
Consider Algorithm \ref{algo:1} under Assumption \ref{assum:1} for solving Problem \ref{prob:1}. Let $\alpha_k$ be a learning rate satisfying the Armijo condition (\ref{armijo}) (whose existence is guaranteed by Proposition \ref{existence_armijo}), let $L_{B_k}$ be the Lipschitz constant of $\nabla f_{B_k}$. Then, the following hold. 
\begin{enumerate}
\item[{\em (i)}] {\em [Counter-example of \citep[Lemma 1]{NEURIPS2019_2557911c}]} There exists Problem \ref{prob:1} such that $\alpha_k$ does not satisfy $\min\{ \frac{2 (1-c)}{L_{B_k}}, \overline{\alpha} \} \leq \alpha_k$, where $\overline{\alpha}$ is an upper bound of $\alpha_k$. 
\item[{\em (ii)}] {\em [Lower bound on learning rate determined by backtracking line search method]} If $\alpha_k$ can be computed by Algorithm \ref{algo:2}, then there exists a lower bound of $\alpha_k$ such that 
$0 < \underline{\alpha} :=  \frac{2 \delta (1-c)}{L} \leq \alpha_k$, where $L$ is the maximum value of $L_i$.
\end{enumerate}
\end{lemma}

\section{Analysis of SGD using Armijo Line Search}
\subsection{Convergence analysis of Algorithm \ref{algo:1}}
\label{subsec:3.1}
Here, we present a convergence analysis of Algorithm \ref{algo:1}. The proof of Theorem \ref{thm:1} is given in Appendix \ref{a2}.

\begin{theorem}
[Upper bound of the squared norm of the full gradient]\label{thm:1} 
Consider the sequence $(\bm{\theta}_k)_{k\in \mathbb{N}}$ generated by Algorithm \ref{algo:1} under Assumption \ref{assum:1} for solving Problem \ref{prob:1} and suppose that the learning rate $\alpha_k \in [\underline{\alpha}, \overline{\alpha}]$ is computed by Algorithm \ref{algo:2}. Then, for all $K \geq 1$, the following hold:
\begin{enumerate}
\item[{\em (i)}]
In the case of $\frac{1}{L_n}\geq\overline{\alpha}$,
\begin{align*}
\min_{k\in [0:K-1]} \mathbb{E}\left[ \| \nabla f(\bm{\theta}_k)\|^2 \right]
\leq 
\underbrace{\overbrace{\frac{2(f(\bm{\theta}_{0}) - f_*)}
{ (2-L_n \overline{\alpha} ) \underline{\alpha}}}^{C_1}
\frac{1}{K}}_{B(\bm{\theta}_0, K)} 
+ 
\underbrace{\overbrace{\frac{L_n \overline{\alpha}^2 \sigma^2}{ (2-L_n \overline{\alpha} ) \underline{\alpha}}}^{C_2} \frac{1}{b}}_{V(\sigma^2, b)},
\end{align*}
where 
$L_n := \frac{1}{n} \sum_{i\in [n]} L_i$, $f_* := \frac{1}{n} \sum_{i\in [n]} f_{i,*}$, $\underline{\alpha} := \frac{2 \delta (1-c)}{L}$, 
$\delta \in (0,1)$, and $c \in (0,1)$.
\item[{\em (ii)}]
Suppose that the random variable $\xi_k$ follows a discrete uniform distribution $\mathrm{DU}_b (n)$.
In the case of $\frac{1}{L_n}< \overline{\alpha}$,
\begin{align*}
\min_{k\in [0:K-1]} \mathbb{E}\left[ \| \nabla f(\bm{\theta}_k)\|^2 \right]
\leq 
\underbrace{\overbrace{\frac{2(f(\bm{\theta}_{0}) - f_*)}
{\tilde{\alpha} - (L_n \overline{\alpha} -1) \overline{\alpha}}}^{C_1}
\frac{1}{K}}_{B(\bm{\theta}_0, K)} 
+ 
\underbrace{\overbrace{\frac{L_n \overline{\alpha}^2 \sigma^2}{\tilde{\alpha} - (L_n \overline{\alpha} -1) \overline{\alpha}}}^{C_2} \frac{1}{b}}_{V(\sigma^2, b)},
\end{align*}
where 
$L_n := \frac{1}{n} \sum_{i\in [n]} L_i$, $f_* := \frac{1}{n} \sum_{i\in [n]} f_{i,*}$, 
$\tilde{\alpha}:=\frac{2 \delta (1-c)}{L_n}$,
$\delta \in (\frac{1}{4},1)$, and $c \in (0,1 - \frac{1}{4\delta})$.
\end{enumerate}
\end{theorem}

Theorem \ref{thm:1} indicates that the upper bound of the minimum value of $\mathbb{E}[ \| \nabla f(\bm{\theta}_k)\|^2 ]$ consists of a bias term $B(\bm{\theta}_0,K)$ and variance term $V(\sigma^2, b)$. When the number of steps $K$ is large and the batch size $b$ is large, $B(\bm{\theta}_0,K)$ and $V(\sigma^2, b)$ become small. Therefore, we need to set $K$ large and $b$ large so that Algorithm \ref{algo:1} will approximate a local minimizer of $f$.

Here, we compare Theorem \ref{thm:1} with the convergence analysis of SGD using a constant learning rate. SGD using a constant learning rate $\alpha \in (0,\frac{2}{L_n})$ satisfies 
\begin{align}\label{sgd_c}
\min_{k\in [0:K-1]} \mathbb{E}\left[ \| \nabla f(\bm{\theta}_k)\|^2 \right]
\leq 
\frac{2 (f(\bm{\theta}_0) - f_*)}{(2 - L_n \alpha) \alpha}
\frac{1}{K}+ 
\frac{L_n \alpha \sigma^2}{2-L_n \alpha}\frac{1}{b}
\end{align}
(The proof of (\ref{sgd_c}) is given in Appendix \ref{a5}). 
We need to set a constant learning rate $\alpha \in (0,\frac{2}{L_n})$ depending on the Lipschitz constant $L_n$ of $\nabla f$. 
However, since computing $L_n$ is NP-hard \citep{NEURIPS2018_d54e99a6}, it is difficult to set $\alpha \in (0,\frac{2}{L_n})$. 
Meanwhile, from Theorem \ref{thm:1}(ii), we need to set $c,\delta \in (0,1)$ in Algorithm \ref{algo:2} such that  
$\delta \in (\frac{1}{4},1)$ and $c \in (0,1 - \frac{1}{4\delta})$
(see Section \ref{sec:4} for the performance of Algorithms \ref{algo:1} and \ref{algo:2} using $\delta = 0.9$ and small parameters $c$).

We also compare Theorem \ref{thm:1} with Theorem 3 in \citep{NEURIPS2019_2557911c}. Theorem 3 in \citep{NEURIPS2019_2557911c} indicates that, under a strong growth condition with a constant $\rho$ (i.e., $\mathbb{E}_i [\|\nabla f_i (\bm{\theta})\|^2] \leq \rho \| \nabla f(\bm{\theta})\|^2$ $(\bm{\theta} \in \mathbb{R}^d)$) and the Armijo condition, SGD satisfies that
\begin{align*}
\min_{k\in [0:K-1]} \mathbb{E}\left[ \| \nabla f(\bm{\theta}_k)\|^2 \right]
\leq \frac{f(\bm{\theta}_0) - f(\bm{\theta}^\star)}{\Delta K},
\end{align*}
where $L$ is the maximum value of the Lipschitz constant $L_i$ of $\nabla f_i$, 
$c > 1 - \frac{L}{\rho L_n}$, $\overline{\alpha} < \frac{2}{\rho L_n}$, $\Delta := (\overline{\alpha} + \frac{2(1-c)}{L}) - \rho (\overline{\alpha} - \frac{2(1-c)}{L} + L_n \overline{\alpha}^2)$, and $\bm{\theta}^\star$ is a local minimizer of $f$. Theorem \ref{thm:1} is a convergence analysis of Algorithm \ref{algo:1} without assuming the strong growth condition. 
Moreover, Theorem \ref{thm:1} shows that using large batch size is appropriate for SGD using the Armijo line search (Algorithm \ref{algo:1}).

\subsection{Steps needed for $\epsilon$--approximation}
To investigate the relationship between the number of steps $K$ needed for nonconvex optimization and the batch size $b$, we consider an $\epsilon$--approximation of Algorithm \ref{algo:1} defined as follows:
\begin{align}\label{epsilon}
\min_{k\in [0:K-1]} \mathbb{E}\left[ \| \nabla f(\bm{\theta}_k)\|^2 \right]
\leq \epsilon^2,
\end{align}
where $\epsilon > 0$ is the precision.

Theorem \ref{thm:1} leads to the following theorem indicating the relationship between $b$ and the values of $K$ that achieves an $\epsilon$--approximation. The proof of Theorem \ref{thm:2} is given in Appendix \ref{a3}.

\begin{theorem}
[Steps needed for nonconvex optimization of SGD using Armijo line search]\label{thm:2}
Suppose that the assumptions in Theorem \ref{thm:1} hold. Define $K \colon \mathbb{R} \to \mathbb{R}$ for all $b > \frac{C_2}{\epsilon^2}$ by 
\begin{align}\label{K}
K(b) = \frac{C_1 b}{\epsilon^2 b - C_2},
\end{align}
where the positive constants $C_1$ and $C_2$ are defined as in Theorem \ref{thm:1}. Then, the following hold:
\begin{enumerate}
\item[{\em (i)}] {\em [Steps needed for nonconvex optimization]} $K$ defined by (\ref{K}) achieves an $\epsilon$--approximation (\ref{epsilon}).
\item[{\em (ii)}] {\em [Properties of the steps]} $K$ defined by (\ref{K}) is monotone decreasing and convex for $b > \frac{C_2}{\epsilon^2}$.
\end{enumerate}
\end{theorem}

Theorem \ref{thm:2} ensures that the number of steps $K$ needed for SGD using the Armijo line search to be an $\epsilon$--approximation is small when the batch size $b$ is large. Therefore, it is useful to set a sufficiently large batch size in the sense of minimizing the steps needed for an $\epsilon$--approximation of SGD using the Armijo line search.


\subsection{Critical batch size minimizing SFO complexity}\label{sec:3.3}
The following theorem shows the existence of a critical batch size for SGD using the Armijo line search. The proof of Theorem \ref{thm:3} is given in Appendix \ref{a4}. 

\begin{theorem}
[Existence of critical batch size for SGD using Armijo line search]\label{thm:3} Suppose that the assumptions in Theorem \ref{thm:1} hold. Define SFO complexity $N \colon \mathbb{R} \to \mathbb{R}$ for the number of steps $K$, defined by (\ref{K}), needed for an $\epsilon$--approximation (\ref{epsilon}) and for a batch size $b > \frac{C_2}{\epsilon^2}$ by 
\begin{align}\label{SFOC}
N(b) = K(b) b = \frac{C_1 b^2}{\epsilon^2 b - C_2},
\end{align}
where the positive constants $C_1$ and $C_2$ are defined as in Theorem \ref{thm:1}{\em(ii)}. Then, the following hold:
\begin{enumerate}
\item[{\em (i)}] {\em [SFO complexity]} $N$ defined by (\ref{SFOC}) is convex for $b > \frac{C_2}{\epsilon^2}$.
\item[{\em (ii)}] {\em [Critical batch size]} There exists a critical batch size 
\begin{align}\label{ucbs}
b^\star = \frac{2 C_2}{\epsilon^2}
= \frac{\sigma^2}{\epsilon^2}  \frac{L_n^2\alpha^2}{ \{2(1-c)\delta - (L_n \overline{\alpha} -1)L_n \overline{\alpha}\}}
\end{align}
such that $b^\star$ minimizes the SFO complexity (\ref{SFOC}).
\end{enumerate}
\end{theorem}

Theorem \ref{thm:3}(ii) indicates that the critical batch size can be obtained from the hyperparameters. Accordingly, we would like to estimate the critical batch size by using equation (\ref{ucbs}). See Appendix \ref{estimation_cbs} for an estimation of the critical batch size. Therefore, the next section numerically examines the relationship between the batch size $b$ and the number of steps $K$ needed for nonconvex optimization and the relationship between $b$ and the SFO complexity $N$ to check if there is a critical batch size $b^\star$ minimizing $N$ and if the critical batch size $b^\star$ can be estimated from our theoretical results.

\section{Numerical Results}
\label{sec:4}
We verified whether numerical results match our theoretical results (Theorems \ref{thm:2} and \ref{thm:3}). 
We also compared the performance of Algorithm \ref{algo:1} with the performances of other optimizers, such as SGD with a constant learning rate (SGD), momentum method (Momentum), Adam, AdamW, and RMSProp. The learning rate and hyperparameters of the five optimizers used in each experiment were determined on the basis of a grid search.

The metrics were the number of steps $K$ and the SFO complexity $N = Kb$ indicating that the training accuracy is higher than a certain score. We used Algorithm \ref{algo:1} with the Armijo-line-search learning rate computed by Algorithm \ref{algo:2} with $\gamma =2$, $\delta = 0.9$, $\overline{\alpha} = 10$ (see \url{https://github.com/IssamLaradji/sls} for the setting of parameters), and various values of $c$. 

\begin{figure*}[htbp]
\begin{tabular}{cc}
\begin{minipage}[t]{0.45\hsize}
\centering
\includegraphics[width=1\textwidth]{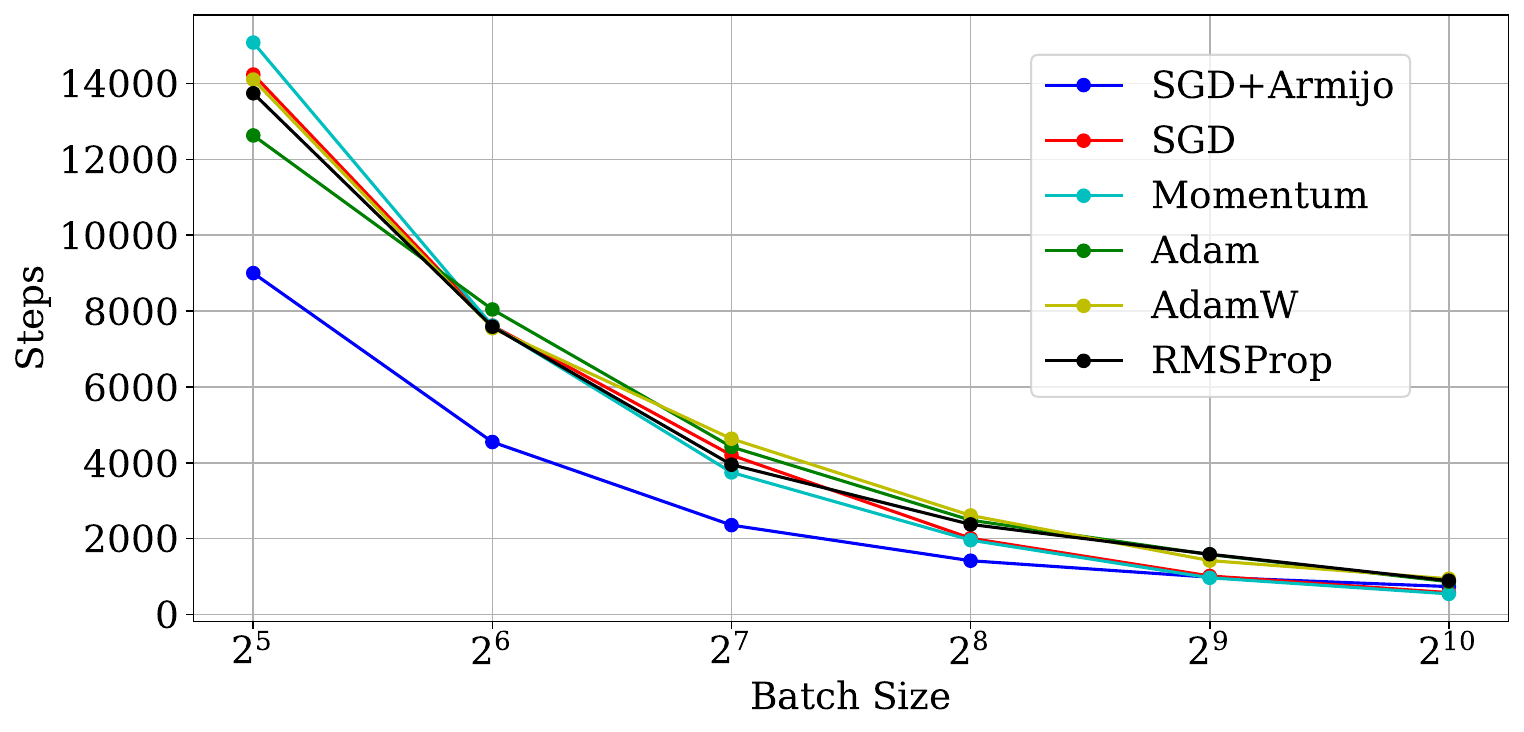}
\caption{Number of steps for Algorithm \ref{algo:1} with $c = 0.20$ and variants of SGD versus batch size needed to train ResNet-34 on CIFAR-10}
\label{fig1_1}
\end{minipage} &
\begin{minipage}[t]{0.45\hsize}
\centering
\includegraphics[width=1\textwidth]{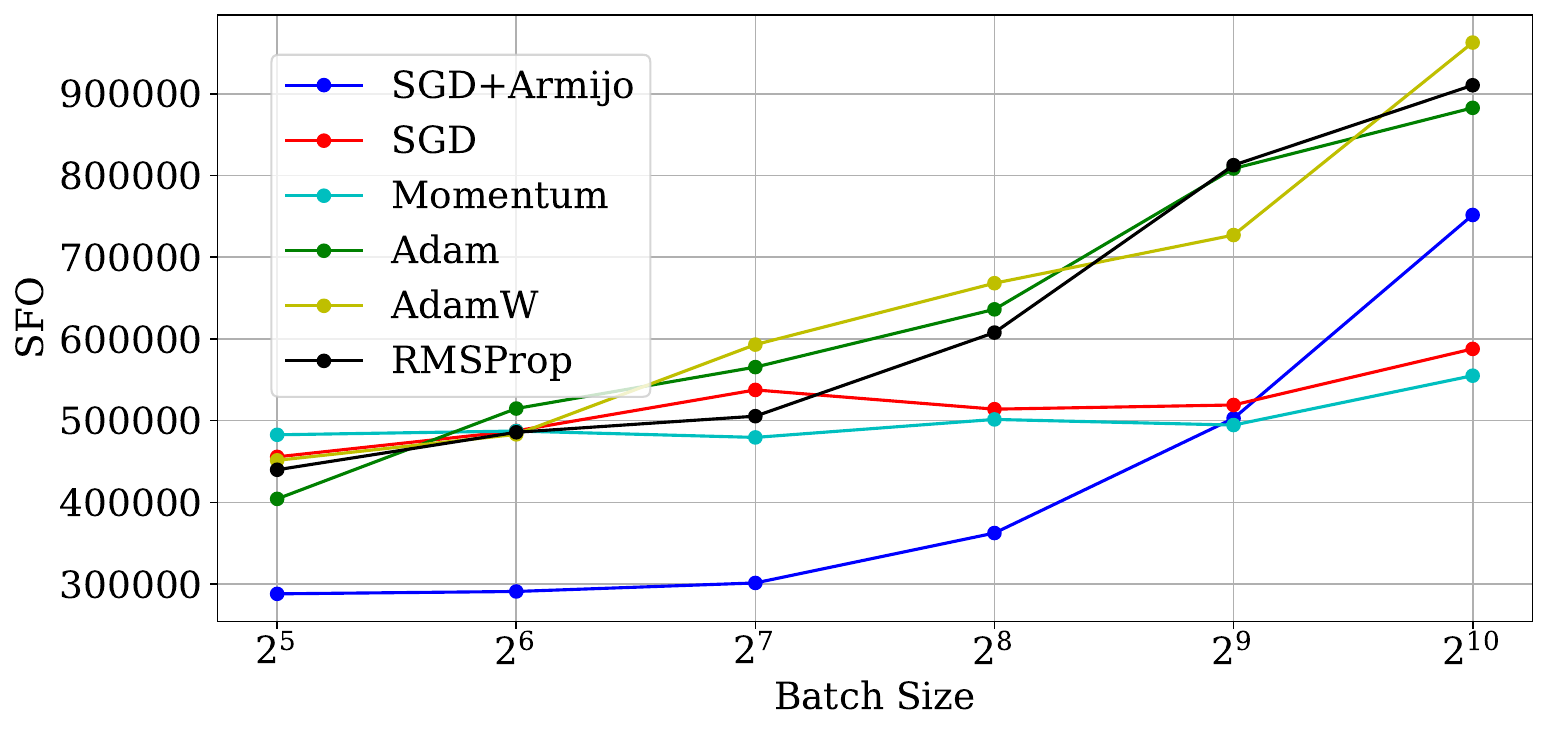}
\caption{SFO complexity for Algorithm \ref{algo:1} with $c = 0.20$ and variants of SGD versus batch size needed to train ResNet-34 on CIFAR-10}
\label{fig2_1}
\end{minipage}
\end{tabular}
\end{figure*}

\begin{figure*}[htbp]
\begin{tabular}{cc}
\begin{minipage}[t]{0.45\hsize}
\centering
\includegraphics[width=1\textwidth]{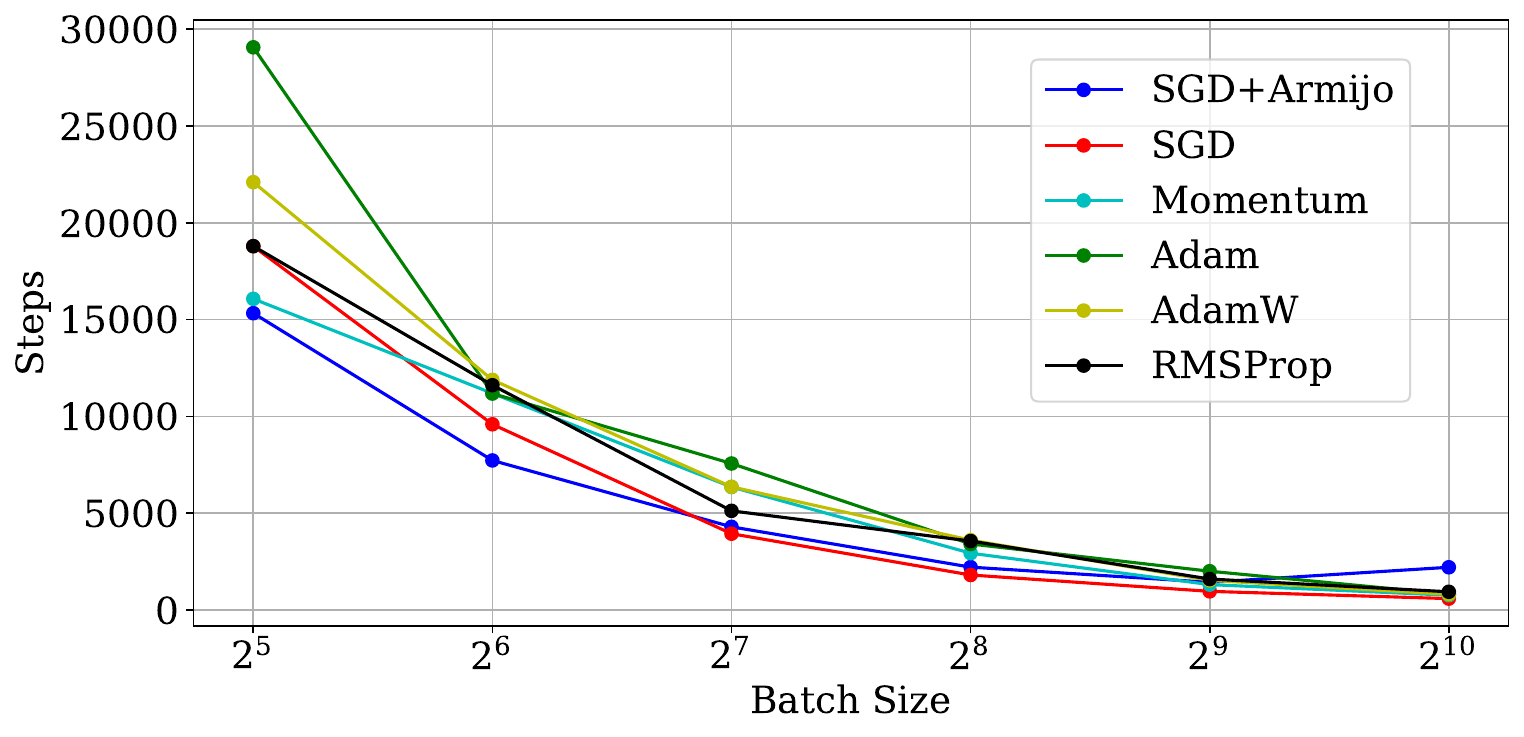}
\caption{Number of steps for Algorithm \ref{algo:1} with $c = 0.25$ and variants of SGD versus batch size needed to train ResNet-34 on CIFAR-100}
\label{fig3_1}
\end{minipage} &
\begin{minipage}[t]{0.45\hsize}
\centering
\includegraphics[width=1\textwidth]{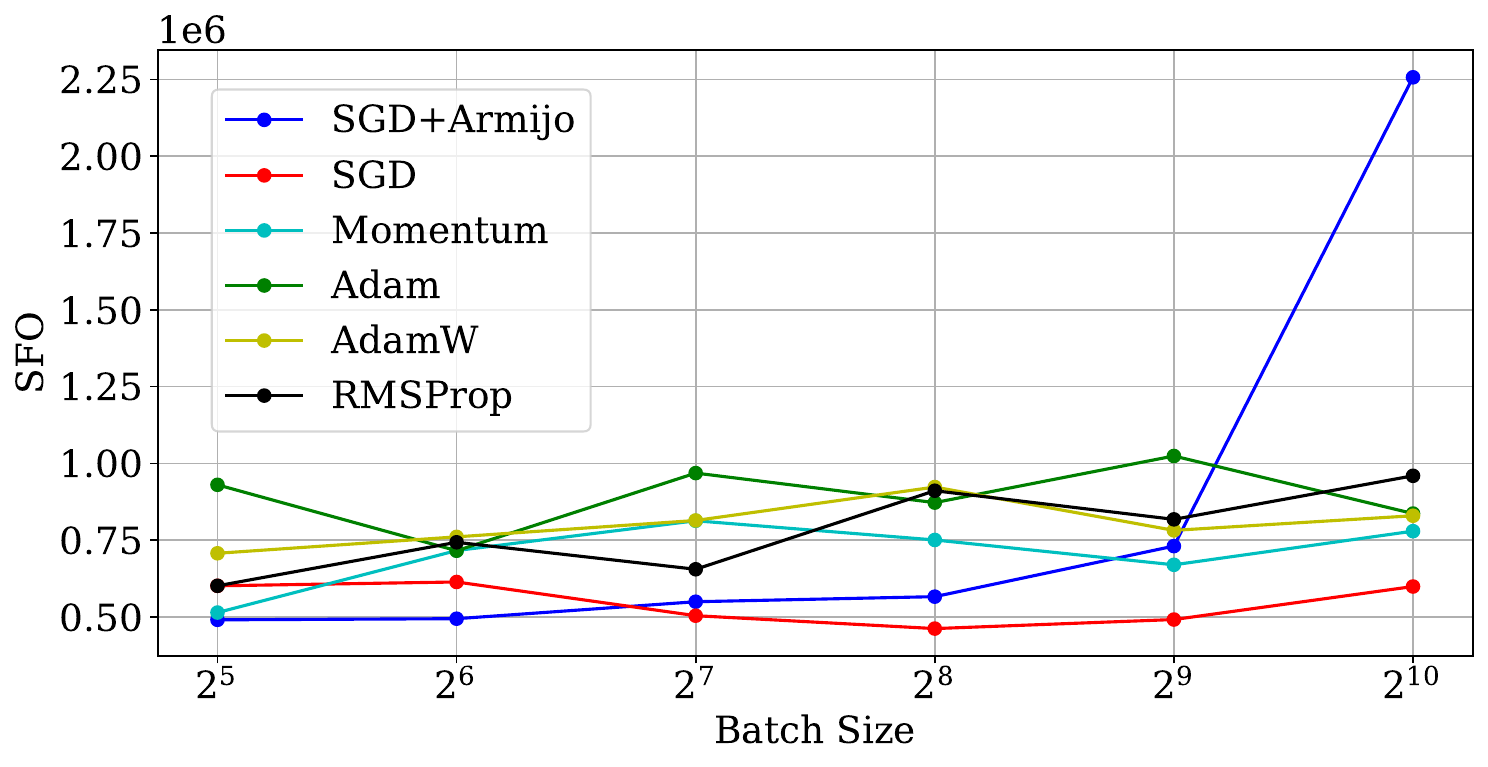}
\caption{SFO complexity for Algorithm \ref{algo:1} with $c = 0.25$ and variants of SGD versus batch size needed to train ResNet-34 on CIFAR-100}
\label{fig4_1}
\end{minipage}
\end{tabular}
\end{figure*}

We trained ResNet-34 on the CIFAR-10 and the CIFAR-100 datasets ($n=50000$) .
See Appendix \ref{add_exp} for numerical results on training MLP on the  MNIST dataset.
Figure \ref{fig1_1} and Figure \ref{fig3_1} plot the number of steps needed for the training accuracy to be more than $0.99$ for Algorithm \ref{algo:1} versus batch size. It can be seen that Algorithm \ref{algo:1} decreases the number of steps as the batch size increases. Figure \ref{fig2_1} and Figure \ref{fig4_1} plot the SFO complexities of Algorithm \ref{algo:1} versus the batch size. It indicates that there are critical batch sizes that minimize the SFO complexities. 

Figure \ref{fig1_1} and Figure \ref{fig2_1} compare the performance of Algorithm \ref{algo:1} with $c = 0.20$ with those of SGD variants. The figures indicate that, when the batch sizes are from $2^5$ to $2^9$, SGD+Armijo (Algorithm \ref{algo:1}) performs better than the other optimizers. In particular, the SFO complexity of SGD+Armijo (Algorithm \ref{algo:1}) using $c = 0.20$ and the critical batch size ($b^\star = 2^5$) is the smallest of  the optimizers for any batch size.

Figure \ref{fig3_1} and Figure \ref{fig4_1} compare the performance of Algorithm \ref{algo:1} with $c = 0.25$ with those of variants of SGD. The figures indicate that, when the batch sizes are from $2^5$ to $2^6$, SGD+Armijo (Algorithm \ref{algo:1}) performs better than the other optimizers. In particular, the SFO complexity of SGD+Armijo (Algorithm \ref{algo:1}) using $c = 0.25$ and the critical batch size ($b^\star = 2^5$) is the smallest of other optimizers for any batch size.

Therefore, we can conclude that Algorithm \ref{algo:1} using the critical batch size $b^\star$ ($\in \{2^5, 2^6\}$) performs better than other optimizers using any batch size in the sense of minimizing the SFO complexities needed to achieve high training accuracies.
 
\section{Conclusion}
This paper presented a convergence analysis of SGD using the Armijo line search for nonconvex optimization. 
We showed that the number of steps needed for nonconvex optimization is monotone decreasing and convex with respect to the batch size; i.e., the steps decrease in number as the batch size increases. We also showed that the SFO complexity needed for nonconvex optimization is convex with respect to the batch size and that there exists a critical batch size at which the SFO complexity is minimized. 
In addition, we gave an equation on the critical batch size and showed that it can be estimated by using some parameters. 
Finally, we provided numerical results that support our theoretical findings. 

\bibliography{biblio}
\bibliographystyle{tmlr}
\newpage
\appendix
\section{Appendix}
\subsection{Proof of Lemma \ref{lem:1}}\label{a1}
(i) Let $k\in \mathbb{N}$ and let $L_{B_k}$ be the Lipschitz constant of $\nabla f_{B_k}$. Lemma 1 in \citep{NEURIPS2019_2557911c} is as follows:
\begin{align}\label{vas}
\begin{split}
&\forall f_{B_k} \colon \mathbb{R}^d \to \mathbb{R} \text{ } 
\forall c \in (0,1) \text{ } 
\forall \bm{\theta}_k \in \mathbb{R}^d \text{ } 
\forall \overline{\alpha} > 0\\
&
\exists \alpha_k \in (0,\overline{\alpha}] \text{ } 
(f_{B_k}(\bm{\theta}_k - \alpha_k \nabla f_{B_k}(\bm{\theta}_k))
\leq 
f_{B_k}(\bm{\theta}_k) - c \alpha_k \|\nabla f_{B_k}(\bm{\theta}_k)\|^2
)\\
&\quad\Rightarrow 
\min \left\{ \frac{2 (1-c)}{L_{B_k}}, \overline{\alpha} \right\} \leq \alpha_k.
\end{split}
\end{align}
The negative proposition of (\ref{vas}) is as follows:
\begin{align}\label{not_vas}
\begin{split}
&\exists f_{B_k} \colon \mathbb{R}^d \to \mathbb{R} \text{ } 
\exists c \in (0,1) \text{ } 
\exists \bm{\theta}_k \in \mathbb{R}^d \text{ } 
\exists \overline{\alpha} > 0\\
&
\exists \alpha_k \in (0,\overline{\alpha}] \text{ } 
(f_{B_k}(\bm{\theta}_k - \alpha_k \nabla f_{B_k}(\bm{\theta}_k))
\leq 
f_{B_k}(\bm{\theta}_k) - c \alpha_k \|\nabla f_{B_k}(\bm{\theta}_k)\|^2
)\\
&\quad\land 
\min \left\{ \frac{2 (1-c)}{L_{B_k}}, \overline{\alpha} \right\} > \alpha_k.
\end{split}
\end{align}
We will prove that (\ref{not_vas}) holds. Let $n = b = 1$, $d=1$, $c = 0.1$, $\overline{\alpha} = 1$, and $f(\theta) = f_{B_k}(\theta) = \theta^2$. From $\nabla f (\theta) = 2\theta$, we have that $L_{B_k} = 2$. Since $\theta^* = 0$ is the global minimizer of $f$, we set $\theta_k \in \mathbb{R}$ such that $\theta_k \neq \theta^*$. The Armijo condition in this case is such that $(\theta_k - 2 \alpha_k \theta_k)^2 \leq \theta_k^2 - c \alpha_k (2 \theta_k)^2$, which is equivalent to $\alpha_k \leq 1 - c = 0.9$. Hence, 
\begin{align*} 
&\exists \alpha_k \in (0,1] \text{ } 
(\alpha_k \leq 0.9) \land 
\left( \min \left\{ 0.9, 1 \right\} > \alpha_k \right)\\
&\Leftrightarrow
\exists \alpha_k \in (0,\overline{\alpha}] \text{ } 
(\alpha_k \leq 1 - c) \land 
\left( \min \left\{ \frac{2 (1-c)}{L_{B_k}}, \overline{\alpha} \right\} > \alpha_k \right)\\
&\Leftrightarrow
\exists \alpha_k \in (0,\overline{\alpha}] \text{ } 
(f_{B_k}({\theta}_k - \alpha_k \nabla f_{B_k}({\theta}_k))
\leq 
f_{B_k}({\theta}_k) - c \alpha_k \|\nabla f_{B_k}({\theta}_k)\|^2
)\\
&\quad\land 
\min \left\{ \frac{2 (1-c)}{L_{B_k}}, \overline{\alpha} \right\} > \alpha_k,
\end{align*} 
which implies that (\ref{not_vas}) holds.

(ii) 
We can show Lemma 2.1(ii) using the proof (Case 2) of Lemma 1 in \citep{galli2023don}.
Since $\frac{\alpha_k}{\delta}$ does not satisfy the Armijo condition (\ref{armijo}), we have that
\begin{align}\label{ineq:1}
f_{B_k}\left(\bm{\theta}_k - \frac{\alpha_k}{\delta} \nabla f_{B_k}(\bm{\theta}_k) \right)
> 
f_{B_k}(\bm{\theta}_k) - c \frac{\alpha_k}{\delta} \|\nabla f_{B_k}(\bm{\theta}_k)\|^2.
\end{align}
The $L_{B_k}$--smoothness of $f_{B_k}$ ensures that the descent lemma is true, i.e.,
\begin{align*}
&f_{B_k} \left(\bm{\theta}_k - \frac{\alpha_k}{\delta} \nabla f_{B_k}(\bm{\theta}_k) \right)\\
&\leq 
f_{B_k}(\bm{\theta}_k)
+ 
\left\langle \nabla f_{B_k}(\bm{\theta}_k), 
\left( \bm{\theta}_k - \frac{\alpha_k}{\delta} \nabla f_{B_k}(\bm{\theta}_k) \right) - \bm{\theta}_k \right\rangle
+ \frac{L_{B_k}}{2} \left\|\left( \bm{\theta}_k - \frac{\alpha_k}{\delta} \nabla f_{B_k}(\bm{\theta}_k) \right) - \bm{\theta}_k \right\|^2,
\end{align*}
which implies that
\begin{align}\label{ineq:2}
f_{B_k} \left(\bm{\theta}_k - \frac{\alpha_k}{\delta} \nabla f_{B_k}(\bm{\theta}_k) \right)
\leq 
f_{B_k}(\bm{\theta}_k)
+ \frac{\alpha_k}{\delta}\left( \frac{L_{B_k} \alpha_k}{2\delta} -1 \right) \|\nabla f_{B_k}(\bm{\theta}_k)\|^2.
\end{align}
Hence, (\ref{ineq:1}) and (\ref{ineq:2}) imply that
\begin{align*}
- c \frac{\alpha_k}{\delta} \|\nabla f_{B_k}(\bm{\theta}_k)\|^2
\leq 
\frac{\alpha_k}{\delta}\left( \frac{L_{B_k} \alpha_k}{2\delta} -1 \right) \|\nabla f_{B_k}(\bm{\theta}_k)\|^2,
\end{align*}
which in turn implies that 
\begin{align*}
\frac{\alpha_k}{\delta}\left( \frac{L_{B_k} \alpha_k}{2\delta} 
- (1-c) \right) \|\nabla f_{B_k}(\bm{\theta}_k)\|^2 \geq 0.
\end{align*}
Accordingly, 
\begin{align*}
\frac{L_{B_k} \alpha_k}{2\delta} 
- (1-c) \geq 0, \text{ i.e., } 
\alpha_k \geq \frac{2 \delta (1-c)}{L_{B_k}} \geq \frac{2 \delta (1-c)}{L}=:\underline{\alpha}.
\end{align*}
\subsection{Proof of Theorem \ref{thm:1}}\label{a2}
The definition of $f(\bm{\theta}) := \frac{1}{n}\sum_{i\in [n]} f_i(\bm{\theta})$ and the $L_i$--smoothness of $f_i$ $(i\in [n])$ imply that, for all $\bm{\theta}_1, \bm{\theta}_2 \in \mathbb{R}^d$, 
\begin{align*}
\| \nabla f(\bm{\theta}_1) - \nabla f(\bm{\theta}_2) \|
\leq 
\frac{1}{n} \sum_{i\in [n]} \left\|\nabla f_i(\bm{\theta}_1) - \nabla f_i(\bm{\theta}_2) \right\|
\leq 
\frac{\sum_{i\in[n]} L_i}{n} \|\bm{\theta}_1 - \bm{\theta}_2\|,
\end{align*}
which in turn implies that $\nabla f$ is Lipschitz continuous with Lipschitz constant $L_n := \frac{1}{n}\sum_{i\in[n]} L_i$. Hence, the descent lemma ensures that, for all $k \in \mathbb{N}$, 
\begin{align*}
f(\bm{\theta}_{k+1}) 
\leq f(\bm{\theta}_{k}) + \langle \nabla f(\bm{\theta}_{k}),
\bm{\theta}_{k+1} - \bm{\theta}_{k} \rangle + \frac{L_n}{2} 
\|\bm{\theta}_{k+1} - \bm{\theta}_{k}\|^2,
\end{align*}
which, together with $\bm{\theta}_{k+1} := \bm{\theta}_{k} - \alpha_k \nabla f_{B_k}(\bm{\theta}_k)$, implies that 
\begin{align}\label{ineq_main}
f(\bm{\theta}_{k+1}) 
\leq f(\bm{\theta}_{k}) - \alpha_k \langle \nabla f(\bm{\theta}_{k}),
\nabla f_{B_k}(\bm{\theta}_k) \rangle + \frac{L_n \alpha_k^2}{2} 
\|\nabla f_{B_k}(\bm{\theta}_k)\|^2.
\end{align}
From $\langle \bm{x},\bm{y} \rangle = \frac{1}{2} (\|\bm{x}\|^2 + \|\bm{y}\|^2 - \|\bm{x}-\bm{y}\|^2)$ $(\bm{x},\bm{y}\in \mathbb{R}^d)$, we have that, for all $k\in\mathbb{N}$,
\begin{align*}
\langle \nabla f(\bm{\theta}_{k}),
\nabla f_{B_k}(\bm{\theta}_k) \rangle
= \frac{1}{2} 
\left(
\|\nabla f(\bm{\theta}_{k})\|^2 + \|\nabla f_{B_k}(\bm{\theta}_k) \|^2
-
\|\nabla f(\bm{\theta}_{k}) - \nabla f_{B_k}(\bm{\theta}_k) \|^2
\right).
\end{align*}
Accordingly, (\ref{ineq_main}) implies that, for all $k \in \mathbb{N}$,
\begin{align*}
f(\bm{\theta}_{k+1}) 
&\leq f(\bm{\theta}_{k}) 
- \frac{\alpha_k}{2}
\left(
\|\nabla f(\bm{\theta}_{k})\|^2 + \|\nabla f_{B_k}(\bm{\theta}_k) \|^2
-
\|\nabla f(\bm{\theta}_{k}) - \nabla f_{B_k}(\bm{\theta}_k) \|^2
\right)
+ \frac{L_n \alpha_k^2}{2} 
\|\nabla f_{B_k}(\bm{\theta}_k)\|^2\\
&=
f(\bm{\theta}_{k})
- \frac{\alpha_k}{2} \|\nabla f(\bm{\theta}_{k})\|^2
+ \frac{1}{2} (L_n \alpha_k - 1)\alpha_k \|\nabla f_{B_k}(\bm{\theta}_k) \|^2
+ \frac{\alpha_k}{2}\|\nabla f(\bm{\theta}_{k}) - \nabla f_{B_k}(\bm{\theta}_k) \|^2.
\end{align*}

(i) We consider the case of $\frac{1}{L_n} \geq \overline{\alpha}$.
The condition $0< \alpha_k \leq \overline{\alpha}$ implies that 
$L_n \alpha_k - 1 \leq L_n \overline{\alpha} - 1$ and $0\geq L_n \overline{\alpha} - 1$.
From $0< \underline{\alpha} = \frac{2 \delta (1-c)}{L} \leq \alpha_k$, we have that, for all $k\in \mathbb{N}$,
\begin{align}\label{ineq_main_2}
f(\bm{\theta}_{k+1}) 
&\leq 
f(\bm{\theta}_{k})
- \frac{\underline{\alpha}}{2} \|\nabla f(\bm{\theta}_{k})\|^2
+ \frac{1}{2} (L_n \overline{\alpha} - 1)\underline{\alpha} \|\nabla f_{B_k}(\bm{\theta}_k) \|^2
+ \frac{\overline{\alpha}}{2}\|\nabla f(\bm{\theta}_{k}) - \nabla f_{B_k}(\bm{\theta}_k) \|^2.
\end{align}
Assumption \ref{assum:1} guarantees that 
\begin{align}\label{e_1_1}
\mathbb{E}_{\xi_k} \left[\nabla f_{B_k} (\bm{\theta}_k) | \bm{\theta}_k \right] 
= \nabla f(\bm{\theta}_k) \text{ and } 
\mathbb{E}_{\xi_k} \left[\| \nabla f_{B_k} (\bm{\theta}_k) - \nabla f(\bm{\theta}_k) \|^2 | \bm{\theta}_k \right] 
\leq \frac{\sigma^2}{b}.
\end{align}
Hence, we have 
\begin{align}\label{e_2}
\begin{split}
&\mathbb{E}_{\xi_k} \left[\|\nabla f_{B_k} (\bm{\theta}_k)\|^2 | \bm{\theta}_k \right]\\ 
&= 
\mathbb{E}_{\xi_k} \left[\|\nabla f_{B_k} (\bm{\theta}_k) - \nabla f(\bm{\theta}_k) + \nabla f(\bm{\theta}_k) \|^2 | \bm{\theta}_k \right]\\
&=
\mathbb{E}_{\xi_k} \left[\|\nabla f_{B_k} (\bm{\theta}_k) - \nabla f(\bm{\theta}_k) \|^2 | \bm{\theta}_k \right]
+ 2 \mathbb{E}_{\xi_k} \left[ \langle \nabla f_{B_k} (\bm{\theta}_k) - \nabla f(\bm{\theta}_k),
\nabla f(\bm{\theta}_k) \rangle | \bm{\theta}_k \right] 
+ 
\mathbb{E}_{\xi_k} \left[\| \nabla f(\bm{\theta}_k) \|^2 | \bm{\theta}_k \right]\\
&\leq \| \nabla f(\bm{\theta}_k) \|^2 + \frac{\sigma^2}{b}.
\end{split}
\end{align}
Inequalities (\ref{ineq_main_2}), (\ref{e_1_1}), and (\ref{e_2}) guarantee that, for all $k\in \mathbb{N}$,
\begin{align}\label{ineq_main_3}
\begin{split}
\mathbb{E}_{\xi_k} \left[f(\bm{\theta}_{k+1}) | \bm{\theta}_k \right]
&\leq 
f(\bm{\theta}_{k})
-\frac{\underline{\alpha}}{2} \|\nabla f(\bm{\theta}_k)\|^2+ \frac{1}{2} (L_n \overline{\alpha} - 1)\underline{\alpha}
\left( \|\nabla f(\bm{\theta}_k) \|^2 + \frac{\sigma^2}{b}\right)
+ \frac{\overline{\alpha} \sigma^2}{2b}\\
&=
f(\bm{\theta}_{k})
-\frac{\underline{\alpha}}{2} \|\nabla f(\bm{\theta}_k)\|^2-  \left\{ \frac{(L_n \overline{\alpha} -1) \underline{\alpha}}{2} \right\} \|\nabla f(\bm{\theta}_{k})\|^2
+ \frac{L_n \overline{\alpha}^2 \sigma^2}{2b}.
\end{split}
\end{align}
Taking the total expectation on both sides of (\ref{ineq_main_3}) thus ensures that, for all $k \in \mathbb{N}$,
\begin{align}\label{total_1}
\frac{\underline{\alpha} - (L_n \overline{\alpha} -1 )\underline{\alpha}}{2}\leq 
\mathbb{E} \left[ f(\bm{\theta}_{k}) - f(\bm{\theta}_{k+1}) \right]
+ \frac{L_n \overline{\alpha}^2 \sigma^2}{2b}.
\end{align}
Let $K \geq 1$. Summing (\ref{total_1}) from $k=0$ to $k=K-1$ ensures that
\begin{align*}
\frac{\underline{\alpha} - (L_n \overline{\alpha} -1 )\underline{\alpha}}{2}
\sum_{k=0}^{K-1} \mathbb{E}\left[ \| \nabla f(\bm{\theta}_k)\|^2 \right]
\leq 
\mathbb{E} \left[f(\bm{\theta}_{0}) - f(\bm{\theta}_{K}) \right] 
+ \frac{L_n \overline{\alpha}^2 \sigma^2 K}{2b},
\end{align*}
which, together with the boundedness of $f$, i.e., $f_* \leq f(\bm{\theta}_k)$, implies that 
\begin{align*}
\frac{\underline{\alpha} - (L_n \overline{\alpha} -1 )\underline{\alpha}}{2}
\sum_{k=0}^{K-1} \mathbb{E}\left[ \| \nabla f(\bm{\theta}_k)\|^2 \right]
\leq 
\mathbb{E} \left[f(\bm{\theta}_{0}) - f_* \right] 
+ \frac{L_n \overline{\alpha}^2 \sigma^2 K}{2b}.
\end{align*}
Accordingly, 
\begin{align*}
\frac{1}{K} \sum_{k=0}^{K-1} \mathbb{E}\left[ \| \nabla f(\bm{\theta}_k)\|^2 \right]
\leq 
\frac{2(f(\bm{\theta}_{0}) - f_*)}{\{\underline{\alpha} - (L_n \overline{\alpha} -1 )\underline{\alpha} \}K} 
+ \frac{L_n \overline{\alpha}^2 \sigma^2}{\{\underline{\alpha} - (L_n \overline{\alpha} -1 )\underline{\alpha}\}b}.
\end{align*}
Moreover, since we have 
\begin{align*}
\min_{k\in [0:K-1]} \mathbb{E}\left[ \| \nabla f(\bm{\theta}_k)\|^2 \right] \leq \frac{1}{K} \sum_{k=0}^{K-1} \mathbb{E}\left[ \| \nabla f(\bm{\theta}_k)\|^2 \right],
\end{align*}
the assertion in Theorem \ref{thm:1}(i) holds.

(ii)
Let us consider the case of $\frac{1}{L_n} < \overline{\alpha}$. 
From $0< \alpha_k \leq \overline{\alpha}$,
we have that, for all $k\in \mathbb{N}$, 
$L_n \alpha_k - 1 \leq L_n \overline{\alpha} - 1$ and $0< L_n \overline{\alpha} - 1$.
For all $k\in \mathbb{N}$,
\begin{align}\label{ineq_main_2_1}
f(\bm{\theta}_{k+1}) 
&\leq 
f(\bm{\theta}_{k})
- \frac{\alpha_k}{2} \|\nabla f(\bm{\theta}_{k})\|^2
+ \frac{1}{2} (L_n \overline{\alpha} - 1)\overline{\alpha} \|\nabla f_{B_k}(\bm{\theta}_k) \|^2
+ \frac{\overline{\alpha}}{2}\|\nabla f(\bm{\theta}_{k}) - \nabla f_{B_k}(\bm{\theta}_k) \|^2.
\end{align}
Inequalities (\ref{e_1_1}), (\ref{e_2}), and 
(\ref{ineq_main_2_1}) guarantee that, for all $k\in \mathbb{N}$,
\begin{align}\label{ineq_main_3_1}
\mathbb{E}_{\xi_k} \left[f(\bm{\theta}_{k+1}) | \bm{\theta}_k \right]
&\leq 
f(\bm{\theta}_{k})
- \mathbb{E}_{\xi_k} \left[\frac{\alpha_k}{2} \|\nabla f(\bm{\theta}_k)\|^2
\Big| \bm{\theta}_k\right] 
+ \frac{1}{2} (L_n \overline{\alpha} - 1)\overline{\alpha} 
\left( \|\nabla f(\bm{\theta}_k) \|^2 + \frac{\sigma^2}{b}\right)
+ \frac{\overline{\alpha} \sigma^2}{2b}.
\end{align}
Since $\xi_k$ and $\bm{\theta}_k(\xi_{k-1})$ are independent, 
we have that 
\begin{align*}
\mathbb{E}_{\xi_k} \left[\frac{\alpha_k}{2} \|\nabla f(\bm{\theta}_k)\|^2
\Big| \bm{\theta}_k\right] 
= 
\mathbb{E}_{\xi_k} \left[\frac{\alpha_k}{2} \|\nabla f(\bm{\theta}_k)\|^2 \right]
= 
\frac{1}{2} \|\nabla f(\bm{\theta}_k)\|^2 \mathbb{E}_{\xi_k} \left[\alpha_k \right].
\end{align*}
Hence, (\ref{ineq_main_3_1}) implies that 
\begin{align}\label{1}
\mathbb{E}_{\xi_k} \left[f(\bm{\theta}_{k+1})\right]
&\leq
f(\bm{\theta}_{k})
-\frac{1}{2}\mathbb{E}_{\xi_k} \left[\alpha_k\right] \|\nabla f(\bm{\theta}_k)\|^2 
+ \frac{(L_n \overline{\alpha} -1) \overline{\alpha}}{2} \|\nabla f(\bm{\theta}_{k})\|^2
+ \frac{L_n \overline{\alpha}^2 \sigma^2}{2b}.
\end{align}
Here, let us assume that $\xi_k \sim \mathrm{DU}_b(n)$.
Then, we have that 
\begin{align*}
\mathbb{E}_{\xi_k}[L_{B_k}] 
= 
\mathbb{E}_{\xi_k} \left[\frac{1}{b} \sum_{i=1}^b L_{\xi_{k,i}} \right]
= 
\frac{1}{b} \sum_{i=1}^b \mathbb{E}_{\xi_{k,i}} \left[ L_{\xi_{k,i}} \right]
= 
\frac{1}{b} \sum_{i=1}^b \sum_{j=1}^n L_j \mathrm{P}(\xi_{k,i} = j)
= 
\frac{1}{b} \sum_{i=1}^b \frac{1}{n} \sum_{j=1}^n L_j
= L_n.
\end{align*}
Moreover, Jensen's inequality implies that 
\begin{align*}
\mathbb{E}_{\xi_k} \left[\alpha_k \right]
\geq 
\mathbb{E}_{\xi_k} \left[\frac{2 \delta (1-c)}{L_{B_k}} \right]
\geq 
\frac{2 \delta (1-c)}{\mathbb{E}_{\xi_k}[L_{B_k}]} 
= 
\frac{2 \delta (1-c)}{L_n}. 
\end{align*}
Hence, (\ref{1}) ensures that, for all $k \in \mathbb{N}$,
\begin{align}\label{ineq:key_1}
\mathbb{E}_{\xi_k} \left[f(\bm{\theta}_{k+1})\right]
&\leq
f(\bm{\theta}_{k})
- \frac{1}{2} \underbrace{\frac{2\delta (1-c)}{L_n}}_{\tilde{\alpha}} \|\nabla f(\bm{\theta}_k)\|^2 
+ \frac{(L_n \overline{\alpha} -1) \overline{\alpha}}{2} \|\nabla f(\bm{\theta}_{k})\|^2
+ \frac{L_n \overline{\alpha}^2 \sigma^2}{2b}.
\end{align}
Taking the total expectation on both sides of (\ref{ineq:key_1}) thus ensures that, for all $k \in \mathbb{N}$,
\begin{align}\label{total}
\frac{\tilde{\alpha} - (L_n \overline{\alpha} -1 )\overline{\alpha}}{2}
\mathbb{E}\left[\|\nabla f(\bm{\theta}_k)\|^2\right]
\leq 
\mathbb{E} \left[ f(\bm{\theta}_{k}) - f(\bm{\theta}_{k+1}) \right]
+ \frac{L_n \overline{\alpha}^2 \sigma^2}{2b}.
\end{align}
Let $K \geq 1$. Summing (\ref{total}) from $k=0$ to $k=K-1$ ensures that
\begin{align*}
\frac{\tilde{\alpha} - (L_n \overline{\alpha} -1 )\overline{\alpha}}{2}
\sum_{k=0}^{K-1} \mathbb{E}\left[ \| \nabla f(\bm{\theta}_k)\|^2 \right]
\leq 
\mathbb{E} \left[f(\bm{\theta}_{0}) - f(\bm{\theta}_{K}) \right] 
+ \frac{L_n \overline{\alpha}^2 \sigma^2 K}{2b},
\end{align*}
which, together with the boundedness of $f$, i.e., $f_* \leq f(\bm{\theta}_k)$, implies that 
\begin{align*}
\frac{\tilde{\alpha} - (L_n \overline{\alpha} -1 )\overline{\alpha}}{2}
\sum_{k=0}^{K-1} \mathbb{E}\left[ \| \nabla f(\bm{\theta}_k)\|^2 \right]
\leq 
\mathbb{E} \left[f(\bm{\theta}_{0}) - f_* \right] 
+ \frac{L_n \overline{\alpha}^2 \sigma^2 K}{2b}.
\end{align*}
Let $\delta \in (\frac{1}{4},1)$, $c \in (0,1 - \frac{1}{4\delta})$,
and 
\begin{align*}
\hat{\alpha}
:= \frac{1 + \sqrt{1 + 8(1-c)\delta}}{2 L_n}
\underset{[1-c < \frac{1}{\delta}]}< \frac{2}{L_n}.
\end{align*}
Then, we have that $4 (1-c) \delta > 1$, 
$1 - \frac{1}{\delta} < 0 < c$, and $1 - c < \frac{1}{\delta}$.
Hence, 
\begin{align*}
2(1-c) \delta < 2 
&\Leftrightarrow 
2(1-c) \delta -1 < 1 
\Leftrightarrow
8 (1-c)\delta
\left \{
2(1-c) \delta -1
\right\} + 1
< 
8 (1-c)\delta + 1 \\
&\Leftrightarrow
\left\{ 4(1-c) \delta -1 \right\}^2 
< 
8 (1-c)\delta + 1
\Leftrightarrow
4(1-c) \delta   < 1 + \sqrt{8 (1-c)\delta  + 1}\\
&\Leftrightarrow 
\hat{\alpha} := \frac{1 + \sqrt{1 + 8(1-c)\delta}}{2 L_n}> \tilde{\alpha} := \frac{2 \delta (1-c)}{L_n}.
\end{align*}
Since $\tilde{\alpha} < \overline{\alpha} < \hat{\alpha}$,  
we have that $\tilde{\alpha} - (L_n \overline{\alpha} -1 )\overline{\alpha} > 0$.
Therefore, we have 
\begin{align*}
\frac{1}{K} \sum_{k=0}^{K-1} \mathbb{E}\left[ \| \nabla f(\bm{\theta}_k)\|^2 \right]
\leq 
\frac{2(f(\bm{\theta}_{0}) - f_*)}{\{\underline{\alpha} - (L_n \overline{\alpha} -1 )\overline{\alpha} \}K} 
+ \frac{L_n \overline{\alpha}^2 \sigma^2}{\{\underline{\alpha} - (L_n \overline{\alpha} -1 )\overline{\alpha}\}b}.
\end{align*}

\subsection{Proof of Theorem \ref{thm:2}}\label{a3}
(i) We have 
$$\frac{C_1}{K} + \frac{C_2}{b} = \epsilon^2$$ 
is equivalent to 
$$K = K(b) = \frac{C_1 b}{\epsilon^2 b - C_2}.$$
Hence, Theorem \ref{thm:1} leads to an $\epsilon$--approximation.

(ii) We have 
\begin{align*}
\frac{\mathrm{d}K(b)}{\mathrm{d}b} 
= \frac{- C_1 C_2}{(\epsilon^2 b - C_2)^2} \leq 0 \text{ and }
\frac{\mathrm{d}^2 K(b)}{\mathrm{d}b^2} 
= \frac{2 C_1 C_2 \epsilon^2}{(\epsilon^2 b - C_2)^3} \geq 0,
\end{align*}
which implies that $K$ is monotone decreasing and convex with respect to $b$.

\subsection{Proof of Theorem \ref{thm:3}}\label{a4}
(i) From 
\begin{align*}
N(b) = \frac{C_1 b^2}{\epsilon^2 b - C_2},
\end{align*}
we have 
\begin{align*}
\frac{\mathrm{d}N(b)}{\mathrm{d}b} 
= \frac{C_1 b (\epsilon^2 b - 2 C_2)}{(\epsilon^2 b - C_2)^2} \text{ and }
\frac{\mathrm{d}^2 N(b)}{\mathrm{d}b^2} 
= \frac{2 C_1 C_2^2}{(\epsilon^2 b - C_2)^3} \geq 0,
\end{align*}
which implies that $N$ is convex with respect to $b$.

(ii) We have 
\begin{align*}
\frac{\mathrm{d}N(b)}{\mathrm{d}b}
\begin{cases}
< 0 &\text{ if } b < b^\star,\\
= 0 &\text{ if } b = b^\star = \frac{2 C_2}{\epsilon^2},\\
> 0 &\text{ if } b > b^\star.
\end{cases}
\end{align*}
Hence, the point $b^\star$ minimizes $N$.

From Theorem \ref{thm:1}(ii), we have that 
\begin{align*}
L_n := \frac{1}{n} \sum_{i\in [n]} L_i 
= \frac{2\delta (1-c)}{\tilde{\alpha}}.
\end{align*}
Hence, 
\begin{align*}
b^\star 
&= \frac{2 C_2}{\epsilon^2}
= \frac{2 L_n \overline{\alpha}^2 \sigma^2}{
\{ \tilde{\alpha} - (L_n \overline{\alpha} -1)\overline{\alpha}\} \epsilon^2}
=
\frac{2 L_n \overline{\alpha}^2 \sigma^2}{
\{ (2 \delta (1-c)/L_n) - (L_n \overline{\alpha} -1)\overline{\alpha}\} \epsilon^2}\\
&=
\frac{\sigma^2}{\epsilon^2}  \frac{L_n^2\alpha^2}{ \{2(1-c)\delta - (L_n \overline{\alpha} -1)L_n \overline{\alpha}\}}
\end{align*}
\subsection{Proof of (\ref{sgd_c})}
\label{a5}
Let $K \geq 1$. From (\ref{ineq_main}) and $\alpha_k := \alpha > 0$, we have that, for all $k\in\mathbb{N}$,
\begin{align*}
f(\bm{\theta}_{k+1}) 
\leq f(\bm{\theta}_{k}) - \alpha \langle \nabla f(\bm{\theta}_{k}),
\nabla f_{B_k}(\bm{\theta}_k) \rangle + \frac{L_n \alpha^2}{2} 
\|\nabla f_{B_k}(\bm{\theta}_k)\|^2.
\end{align*}
Hence, (\ref{e_1_1}) and (\ref{e_2}) ensure that, for all $k\in\mathbb{N}$,
\begin{align*}
\mathbb{E}\left[f(\bm{\theta}_{k+1}) \right] 
\leq 
\mathbb{E}\left[f(\bm{\theta}_{k}) \right] - \alpha \mathbb{E}\left[ \| \nabla f(\bm{\theta}_{k})\|^2 \right] + \frac{L_n \alpha^2}{2} 
\left( \mathbb{E}\left[ \|\nabla f(\bm{\theta}_k)\|^2 \right] + \frac{\sigma^2}{b} \right),
\end{align*}
which implies that, for all $k \in \mathbb{N}$, 
\begin{align*}
\alpha \left( 1- \frac{L_n \alpha}{2} \right) \mathbb{E}\left[ \|\nabla f(\bm{\theta}_k)\|^2 \right]
\leq \mathbb{E} \left[ f(\bm{\theta}_{k}) - f(\bm{\theta}_{k+1}) \right]
+ \frac{L_n \alpha^2 \sigma^2}{2b}.
\end{align*}
Summing the above inequalities from $k=0$ to $k=K-1$ ensures that 
\begin{align*}
\alpha \left( 1- \frac{L_n \alpha}{2} \right) 
\sum_{k=0}^{K-1} \mathbb{E}\left[ \|\nabla f(\bm{\theta}_k)\|^2 \right]
\leq
\mathbb{E} \left[ f(\bm{\theta}_{0}) - f(\bm{\theta}_{K}) \right]
+ \frac{L_n \alpha^2 \sigma^2 K}{2b}.
\end{align*}
Since $f$ is bounded below by $f_* := \frac{1}{n} \sum_{i\in [n]} f_{i,*}$, we have 
\begin{align*}
\min_{k\in [0:K-1]} \mathbb{E}\left[ \|\nabla f(\bm{\theta}_k)\|^2 \right]
\leq
\frac{1}{K} \sum_{k=0}^{K-1} \mathbb{E}\left[ \|\nabla f(\bm{\theta}_k)\|^2 \right]
\leq
\frac{2 \mathbb{E} \left[ f(\bm{\theta}_{0}) - f_* \right]}{\alpha (2 - L_n \alpha)K}
+ 
\frac{L_n \alpha \sigma^2}{(2 - L_n \alpha)b}.
\end{align*}

\subsection{Additional experiments}\label{add_exp}
\begin{figure*}[htbp]
\begin{tabular}{cc}
\begin{minipage}[t]{0.45\hsize}
\centering
\includegraphics[width=1\textwidth]{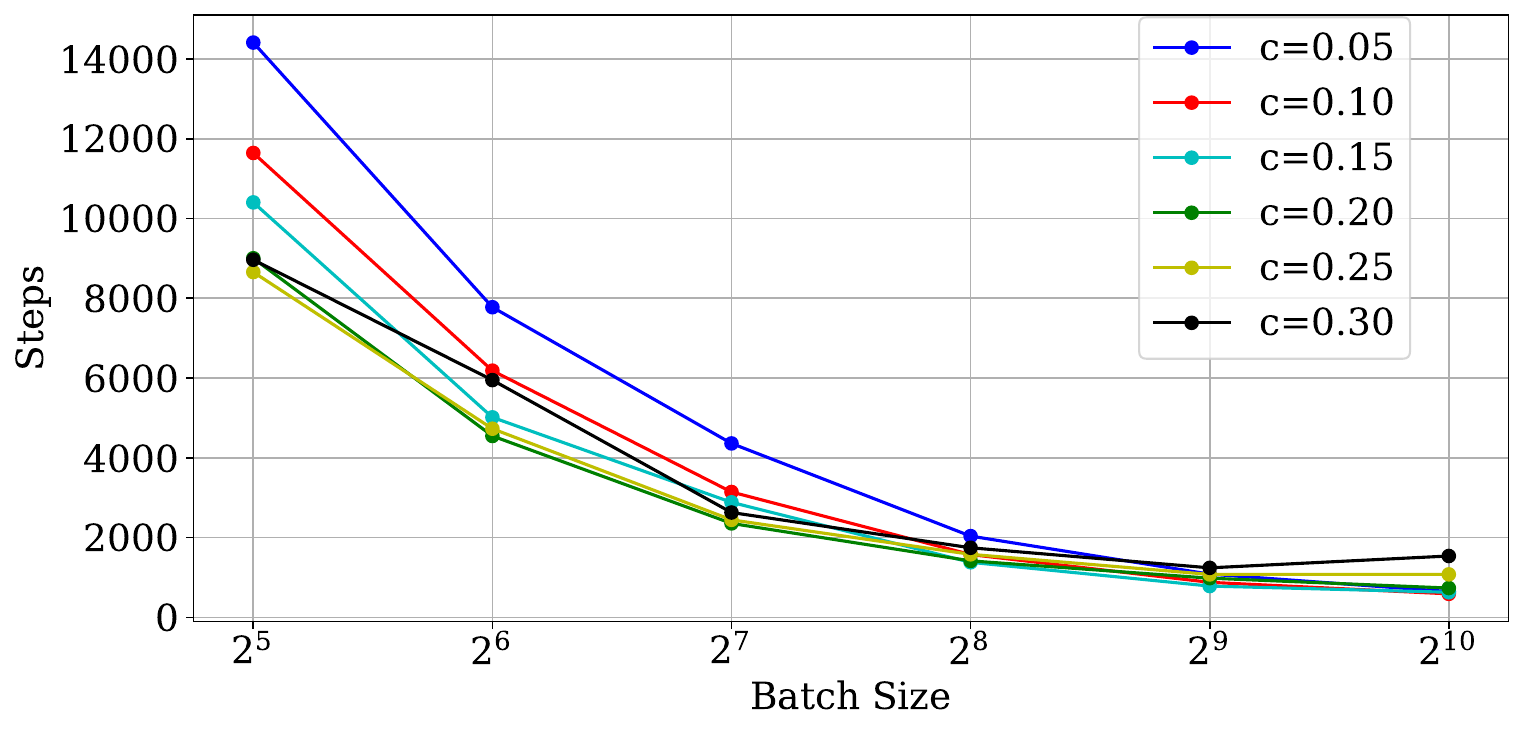}
\caption{Number of steps for Algorithm \ref{algo:1} versus batch size needed to train ResNet-34 on CIFAR-10}
\label{fig1}
\end{minipage} &
\begin{minipage}[t]{0.45\hsize}
\centering
\includegraphics[width=1\textwidth]{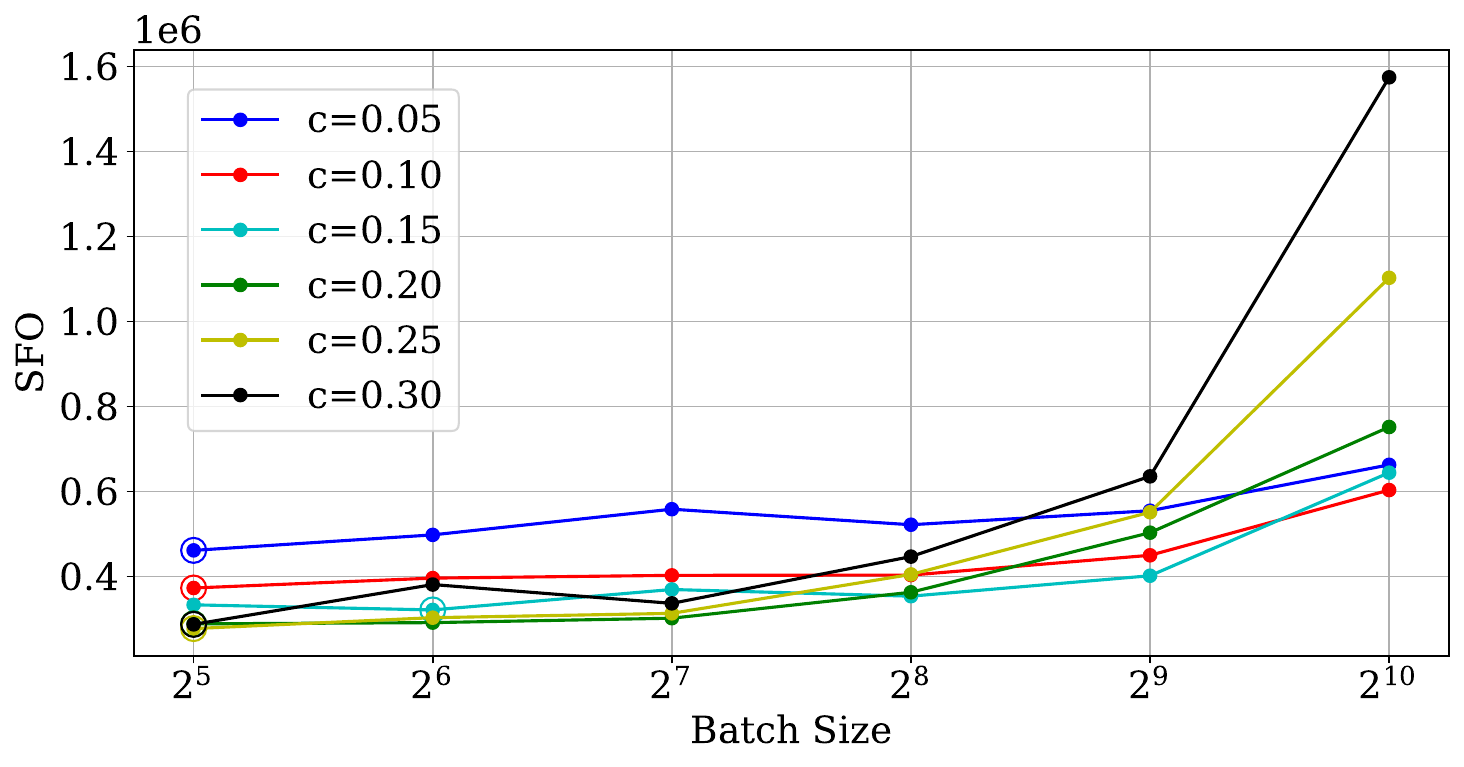}
\caption{SFO complexity for Algorithm \ref{algo:1} versus batch size needed to train ResNet-34 on CIFAR-10 (The double-circle symbol denotes the measured critical batch size)}
\label{fig2}
\end{minipage}
\end{tabular}
\end{figure*}

\begin{figure*}[htbp]
\begin{tabular}{cc}
\begin{minipage}[t]{0.45\hsize}
\centering
\includegraphics[width=1\textwidth]{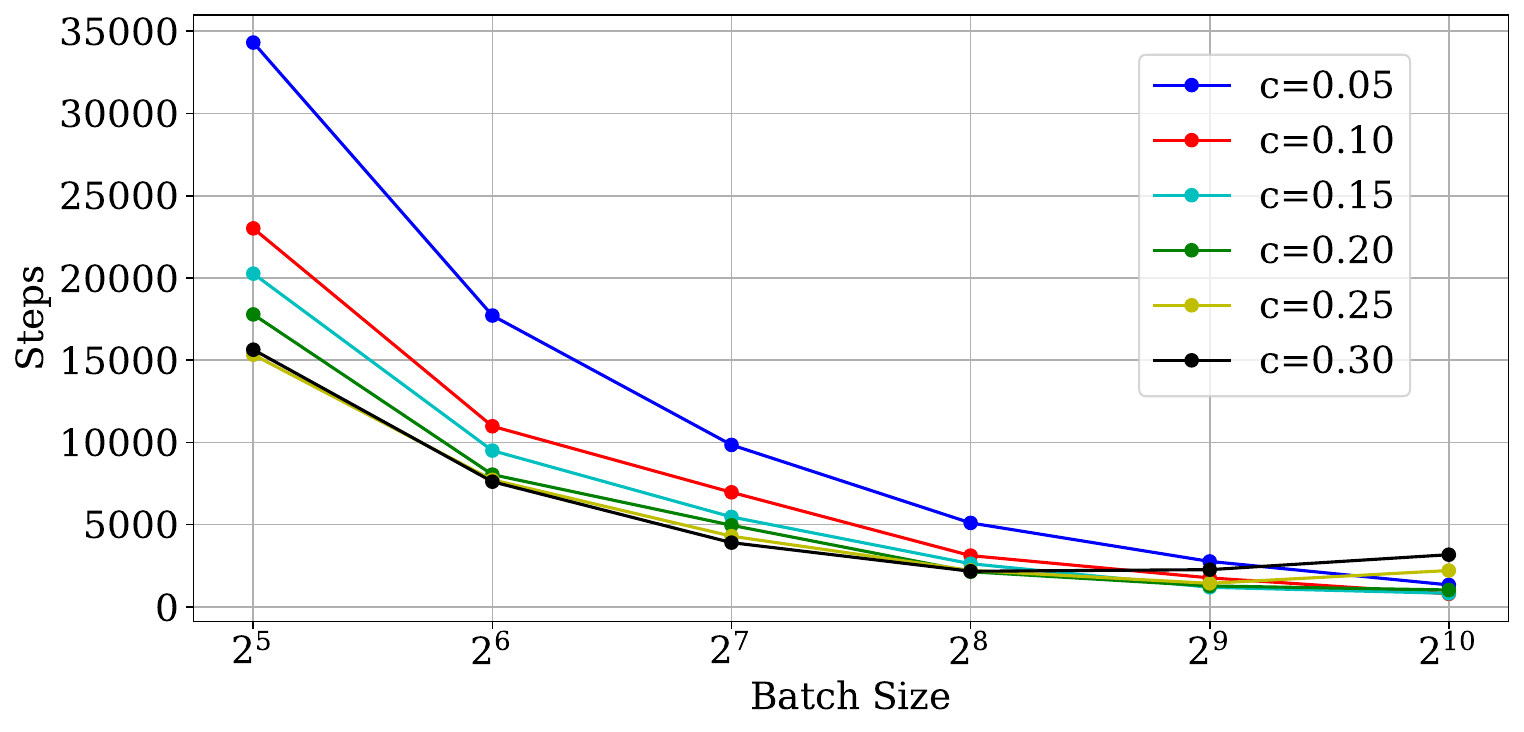}
\caption{Number of steps for Algorithm \ref{algo:1} versus batch size needed to train ResNet-34 on CIFAR-100}
\label{fig3}
\end{minipage} &
\begin{minipage}[t]{0.45\hsize}
\centering
\includegraphics[width=1\textwidth]{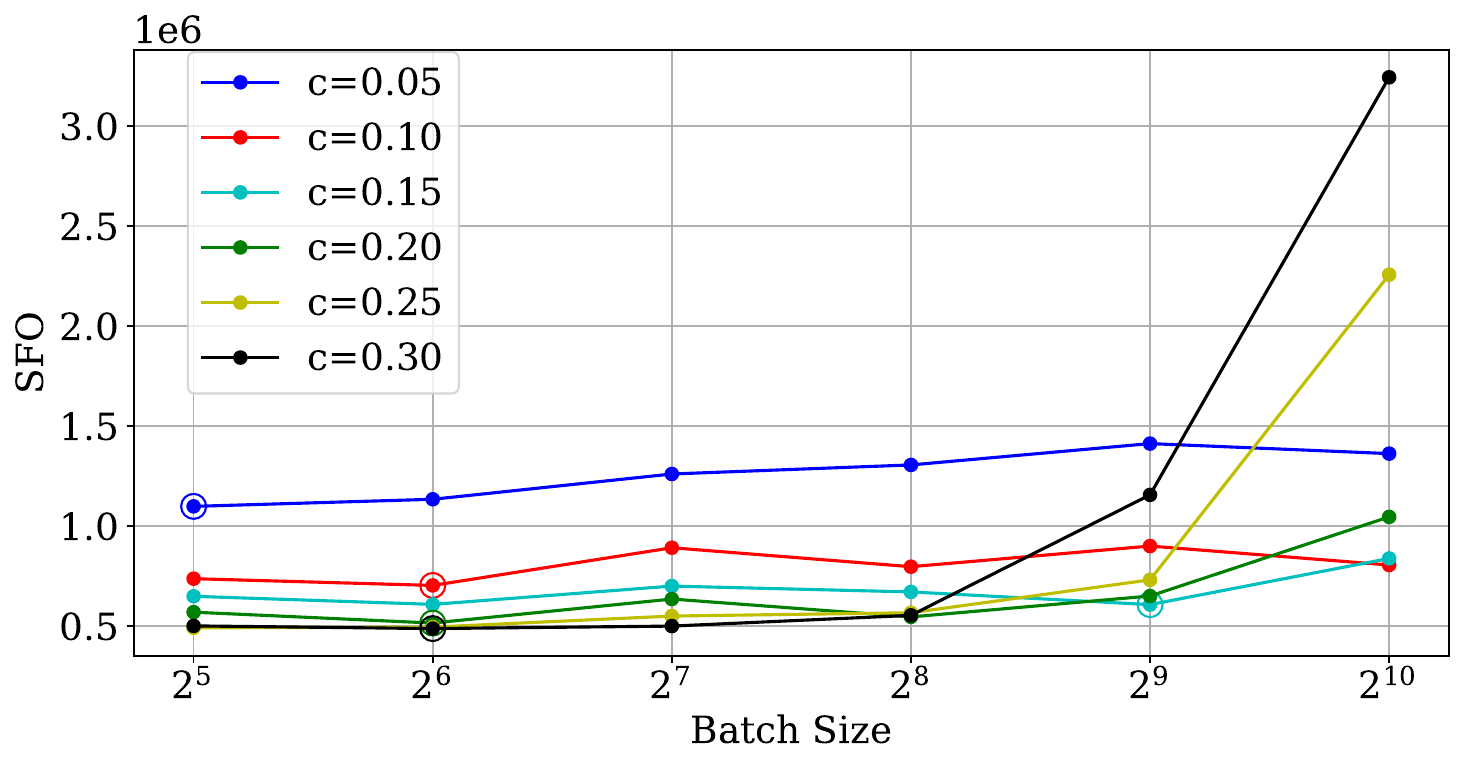}
\caption{SFO complexity for Algorithm \ref{algo:1} versus batch size needed to train ResNet-34 on CIFAR-100 (The double-circle symbol denotes the measured critical batch size)}
\label{fig4}
\end{minipage}
\end{tabular}
\end{figure*}

We trained ResNet-34 on the CIFAR-10 and CIFAR-100 datasets. Figure \ref{fig1} and Figure \ref{fig3} plot the number of steps needed for the training accuracy to be more than $0.99$ versus batch size for Algorithm \ref{algo:1}. It can be seen that Algorithm \ref{algo:1} decreases the number of steps as the batch size increases. Figure \ref{fig2} and Figure \ref{fig4} plot the SFO complexities of Algorithm \ref{algo:1} versus the batch size. They indicate that there are critical batch sizes that minimize the SFO complexities.

\begin{figure*}[htbp]
\begin{tabular}{cc}
\begin{minipage}[t]{0.45\hsize}
\centering
\includegraphics[width=1\textwidth]{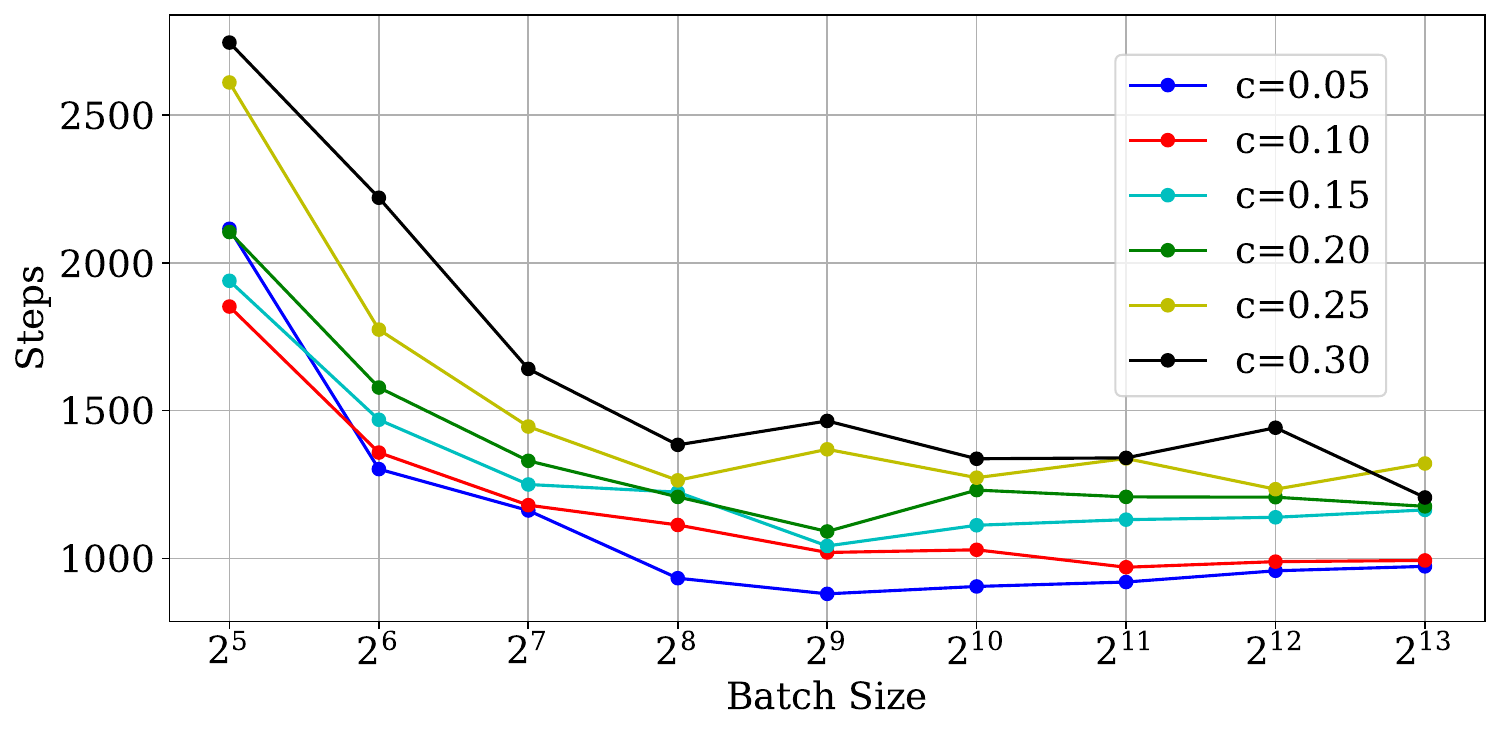}
\caption{Number of steps for Algorithm \ref{algo:1} versus batch size needed to train MLP on MNIST}
\label{fig5}
\end{minipage} &
\begin{minipage}[t]{0.45\hsize}
\centering
\includegraphics[width=1\textwidth]{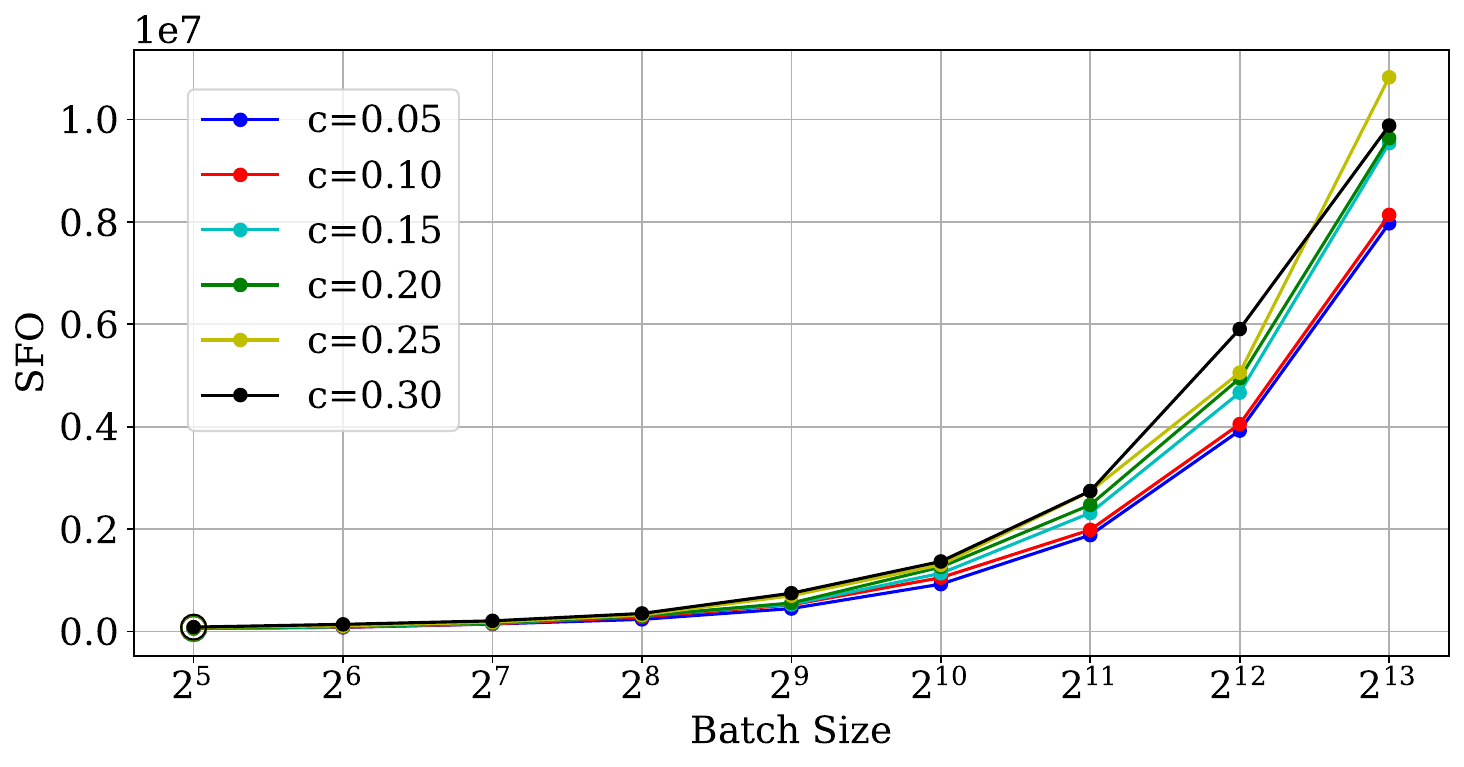}
\caption{SFO complexity for Algorithm \ref{algo:1} versus batch size needed to train MLP on MNIST (The double-circle symbol denotes the measured critical batch size)}
\label{fig6}
\end{minipage}
\end{tabular}
\end{figure*}

We trained a two-hidden-layer MLP with widths of 512 and 256 on the MNIST dataset ($n = 60000$). Figure \ref{fig5} plots the number of steps needed for the training accuracy to be more than $0.97$ for Algorithm \ref{algo:1} versus the batch size, and Figure \ref{fig6} plots the SFO complexities of Algorithm \ref{algo:1} versus the batch size. As in Figures \ref{fig1}, \ref{fig2}, \ref{fig3}, and \ref{fig4}, these figures show that Algorithm \ref{algo:1} decreases the number of steps as the batch size increases and there are critical batch sizes that minimize the SFO complexities.

\begin{figure*}[htbp]
\begin{tabular}{cc}
\begin{minipage}[t]{0.45\hsize}
\centering
\includegraphics[width=1\textwidth]{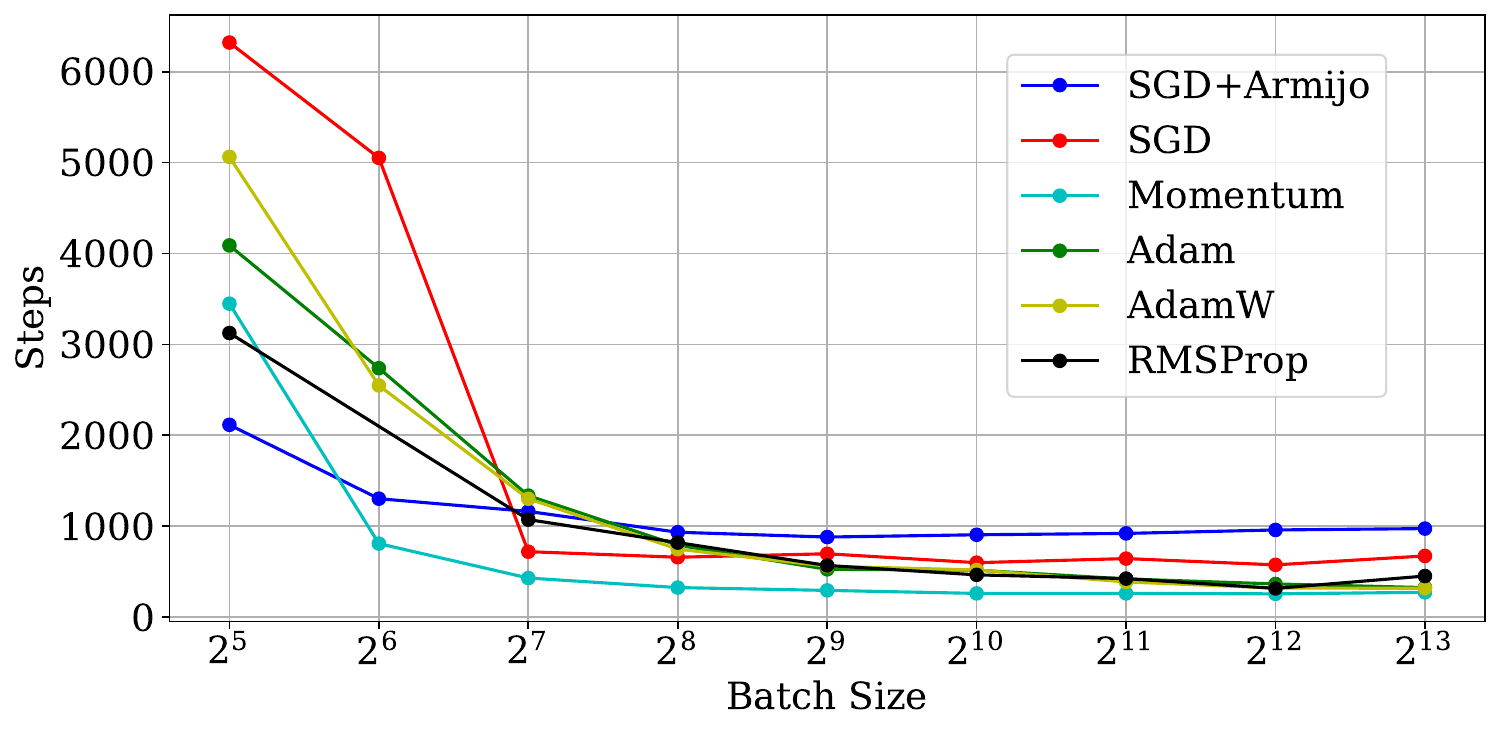}
\caption{Number of steps for Algorithm \ref{algo:1} with $c = 0.05$ and variants of SGD versus batch size needed to train MLP on MNIST}
\label{fig5_1}
\end{minipage} &
\begin{minipage}[t]{0.45\hsize}
\centering
\includegraphics[width=1\textwidth]{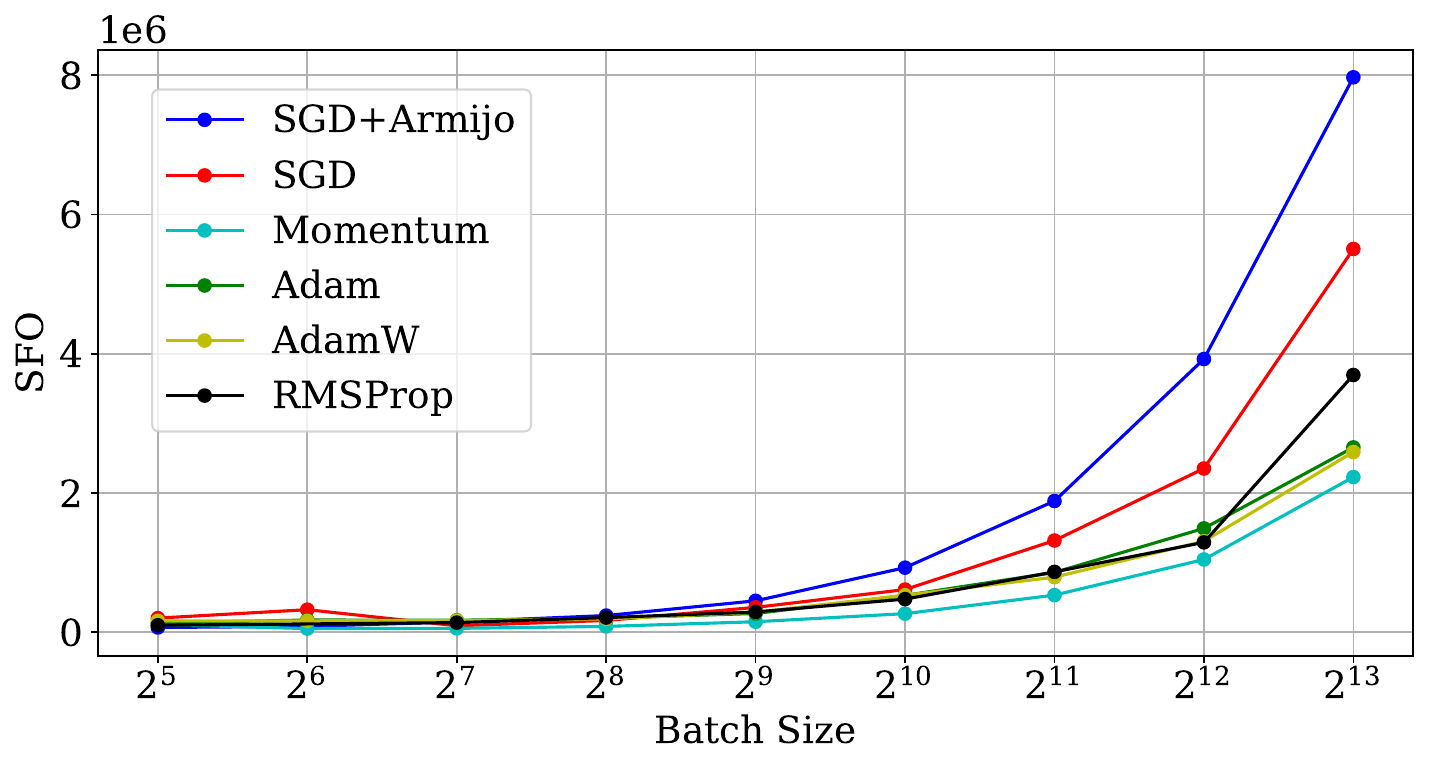}
\caption{SFO complexity for Algorithm \ref{algo:1} with $c = 0.05$ and variants of SGD versus batch size needed to train MLP on MNIST}
\label{fig6_1}
\end{minipage}
\end{tabular}
\end{figure*}

Figures \ref{fig5_1} and \ref{fig6_1} compare the performance of Algorithm \ref{algo:1} with $c = 0.05$ with those of variants of SGD. The figures indicate that SGD+Armijo (Algorithm \ref{algo:1}) using $c = 0.05$ and the critical batch size $(b^\star = 2^5)$ performs better than the other optimizers in the sense of minimizing the SFO complexity. However, as was seen in Figure \ref{fig4_1}, Figure \ref{fig6_1} indicates that the SFO complexity of SGD+Armijo (Algorithm \ref{algo:1}) increases once the batch size exceeds the critical value. 

\begin{figure*}[htbp]
\begin{tabular}{cc}
\begin{minipage}[t]{0.45\hsize}
\centering
\includegraphics[width=1\textwidth]{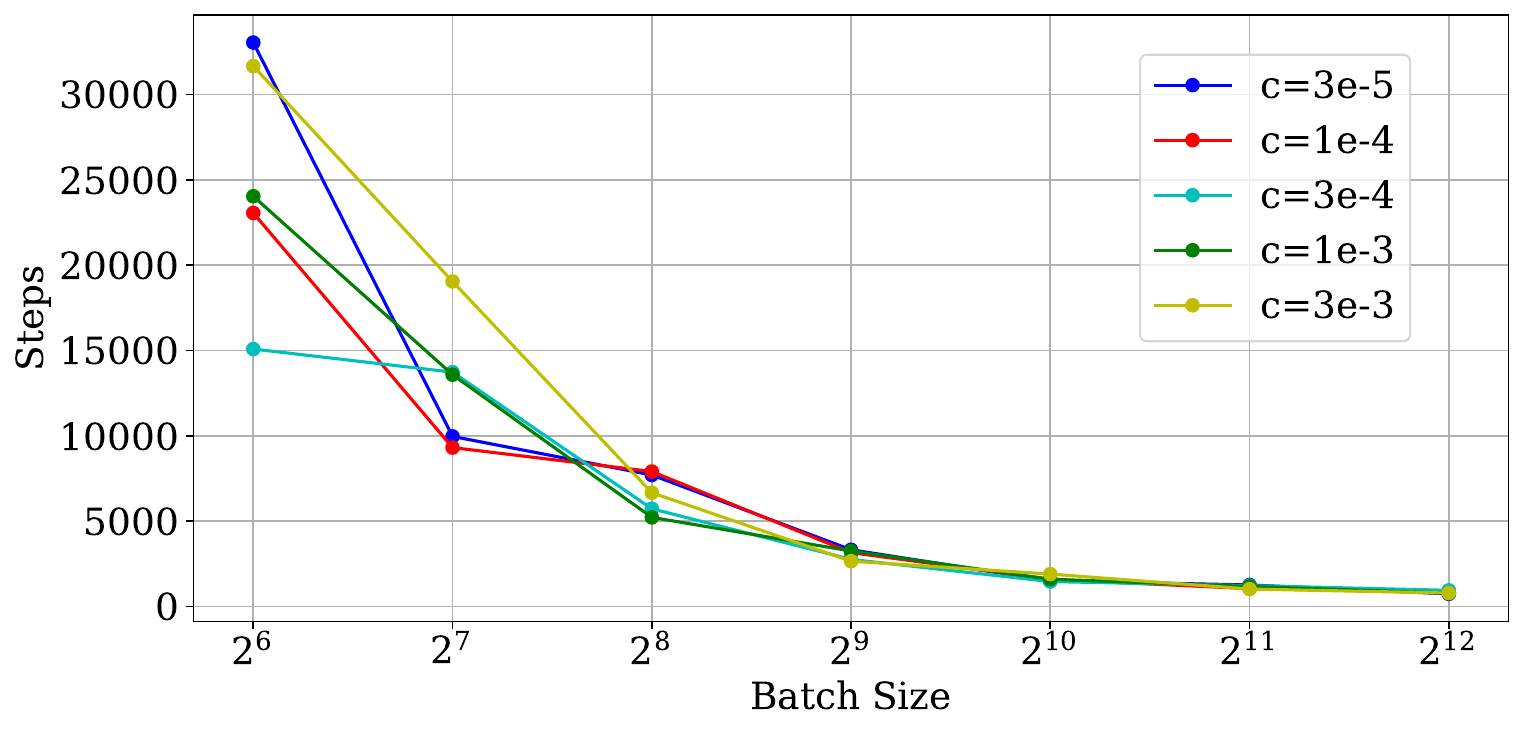}
\caption{Number of steps for Algorithm \ref{algo:1} versus batch size needed to train Wide ResNet-50-2 on CIFAR-10}
\label{fig13}
\end{minipage} &
\begin{minipage}[t]{0.45\hsize}
\centering
\includegraphics[width=1\textwidth]{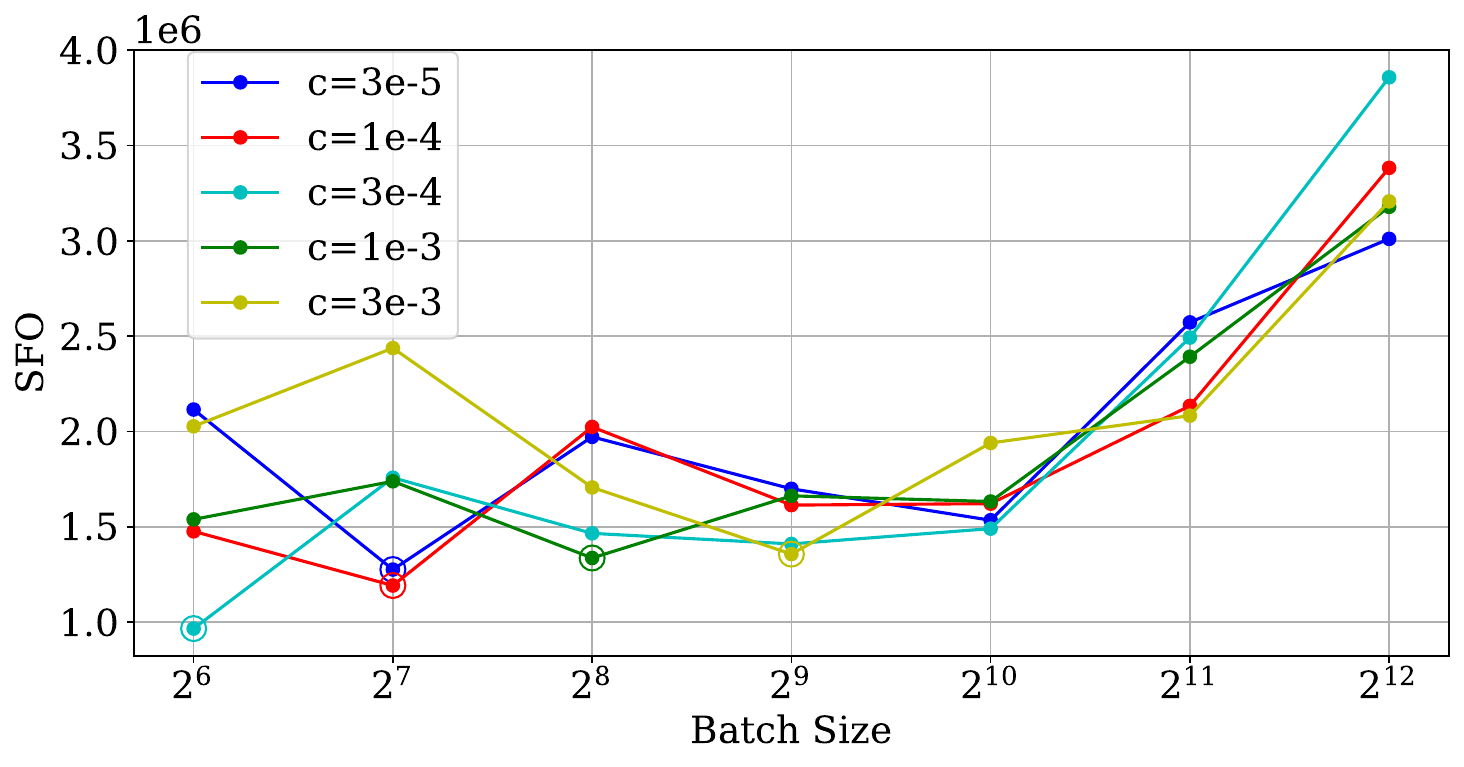}
\caption{SFO complexity for Algorithm \ref{algo:1} versus batch size needed to train Wide ResNet-50-2 on CIFAR-10 (The double-circle symbol denotes the measured critical batch size)}
\label{fig14}
\end{minipage}
\end{tabular}
\end{figure*}

We trained Wide ResNet-50-2 \citep{zagoruyko2017wide} on the CIFAR-10 dataset ($n=50000$). Figure \ref{fig13} plots the number of steps needed for the training accuracy to be more than $0.99$ for Algorithm \ref{algo:1} versus the batch size. It can be seen that Algorithm \ref{algo:1} decreases the number of steps as the batch size increases. Figure \ref{fig14} plots the SFO complexities of Algorithm \ref{algo:1} versus the batch size. It indicates that there are critical batch sizes that minimize the SFO complexities.

\begin{figure*}[htbp]
\begin{tabular}{cc}
\begin{minipage}[t]{0.45\hsize}
\centering
\includegraphics[width=1\textwidth]{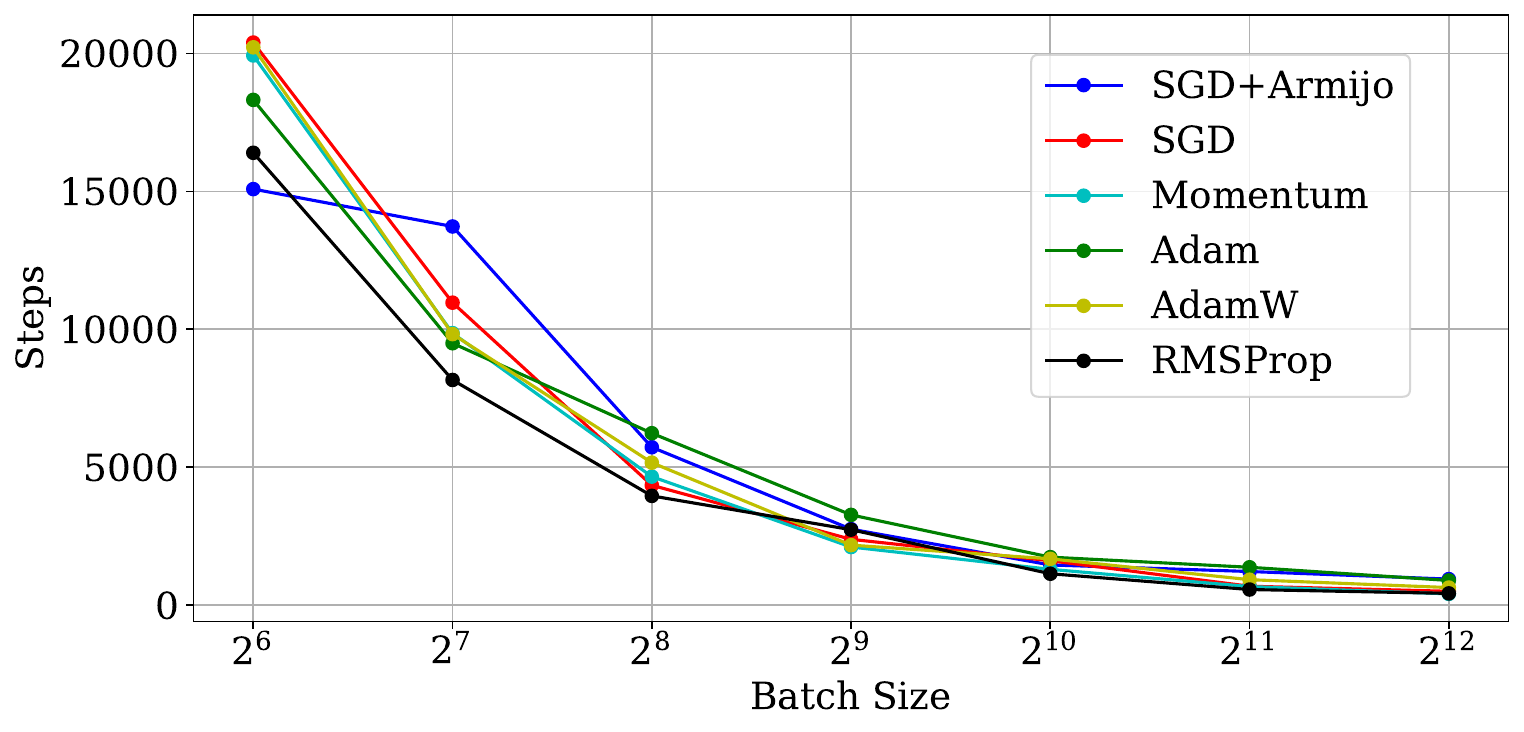}
\caption{Number of steps for Algorithm \ref{algo:1} with $c = 3 \times 10^{-4}$ and variants of SGD versus batch size needed to train Wide ResNet-50-2 on CIFAR-10}
\label{fig15}
\end{minipage} &
\begin{minipage}[t]{0.45\hsize}
\centering
\includegraphics[width=1\textwidth]{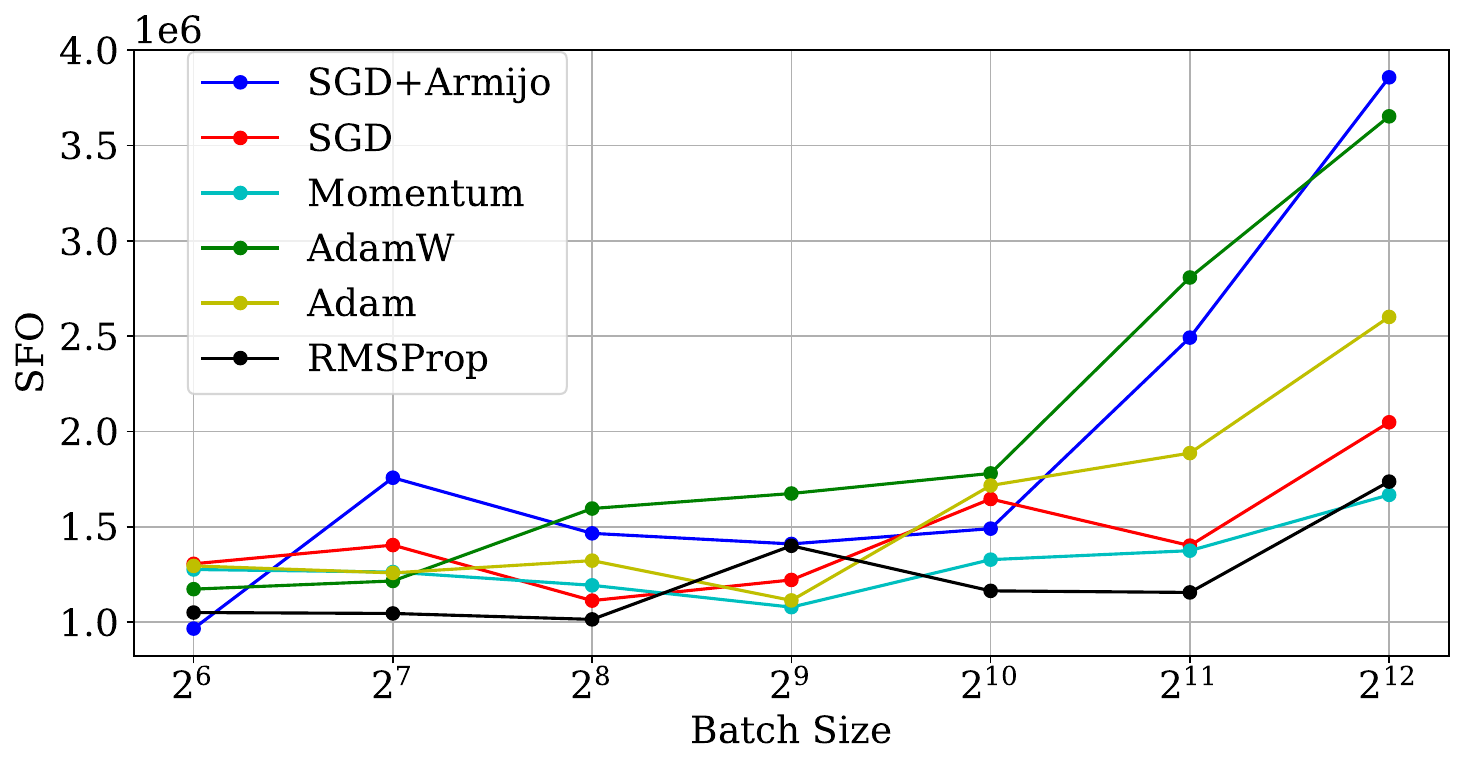}
\caption{SFO complexity for Algorithm \ref{algo:1} with $c = 3 \times 10^{-4}$ and variants of SGD versus batch size needed to train Wide ResNet-50-2 on CIFAR-10}
\label{fig16}
\end{minipage}
\end{tabular}
\end{figure*}
Figures \ref{fig15} and \ref{fig16} compare the performance of Algorithm \ref{algo:1} with $c = 3 \times 10^{-4}$ with those of variants of SGD. The figures indicate that SGD+Armijo (Algorithm \ref{algo:1}) using $c = 3 \times 10^{-4}$ and the critical batch size $(b^\star = 2^6)$ performs better than the other optimizers in the sense of minimizing the SFO complexity. However, as seen in Figures \ref{fig4_1} and \ref{fig6_1}, Figure \ref{fig16} indicates that the SFO complexity of SGD+Armijo (Algorithm \ref{algo:1}) increases once the batch size exceeds the critical value.

\subsection{Estimation of critical batch size}\label{estimation_cbs}
We estimated the critical batch size by using Theorem \ref{thm:3}(iii) and the ideas presented in \citep{iiduka2022critical} and \citep{pmlr-v202-sato23b}. We used Algorithm \ref{algo:1} with $c=0.05$ for training ResNet-34 on the CIFAR-100 dataset (Figures \ref{fig3} and \ref{fig4}). Theorem \ref{thm:3}(iii) indicates that the equation of the critical batch size involves the unknown value $\sigma^2$. We checked that the Armijo-line-search learning rates for Algorithm \ref{algo:1} with $c=0.05$ are about $10$ (see also \citep[Figure 5 (Left)]{NEURIPS2019_2557911c}). Hence, we used $\overline{\alpha} \approx 10$. We estimated the unknown values $X = \frac{\sigma^2}{\epsilon^2}$ and $L_n$ in equation (\ref{ucbs}) of the critical batch size by using $\delta = 0.9$, $b^\star = 2^5$ (see Figure \ref{fig4}), and $\overline{\alpha} \approx 10$ as follows:
\begin{align*} 
b^\star =
\frac{\sigma^2}{\epsilon^2}  \frac{L_n^2\overline{\alpha}^2}{ \{2(1-c)\delta - (L_n \overline{\alpha} -1)L_n \overline{\alpha}\}}\\
X=\frac{32(1.71-100L_n^2+10L_n)}{100L_n^2}
\end{align*}
On the other hand, setting $c=0.30$ and $b^\star = 2^6$ (see Figure \ref{fig4}) gives
\begin{align*} 
b^\star =
\frac{\sigma^2}{\epsilon^2}  \frac{L_n^2\overline{\alpha}^2}{ \{2(1-c)\delta - (L_n \overline{\alpha} -1)L_n \overline{\alpha}\}}\\
X=\frac{64(1.26-100L_n^2+10L_n)}{100L_n^2}
\end{align*}
Let us estimate the critical batch size using $X = 12.3$, $L_n=0.153$, and Theorem \ref{thm:3}(iii). For example, when Algorithm \ref{algo:1} with $c=0.25$ is used to train ResNet-34 on the CIFAR-100 dataset, the equation of the critical batch size is 
\begin{align*}
\frac{L_n^2\alpha^2}{ \{2(1-c)\delta - (L_n \overline{\alpha} -1)L_n \overline{\alpha}\}}X \approx 53 \approx 2^6 = b^\star,
\end{align*}
which implies that the estimated critical batch size $53$ is close to the measured critical batch size $b^\star = 2^6 = 64$ in Figure \ref{fig4}. 
\end{document}